\documentclass[journal,peerreview,nodraftcls]{IEEEtran}
\usepackage{graphicx,subfig,amsthm,amsmath,latexsym,amssymb,times,bm}
\usepackage{float,epsfig,multirow,rotating,times,verbatim,wrapfig,booktabs}
\usepackage{color,xr,array,hyperref,chngcntr}

  \usepackage{cite}

\input{a00-preamble}
\usepackage{amsmath}

\newtheorem{thm}{Theorem}

\newtheorem{result}[thm]{Result}

\newcommand{\R}{\mathbb{R}}

\newcommand{\citep}{\cite}
\newcommand{\citet}{\cite}
\newcommand{\citeyear}{\cite}
\newcommand{\citeauthor}{\cite}

\begin{document}

\title{TiK-means: Transformation-infused $K$-means clustering for skewed groups}%
\author{Nicholas~S.~Berry~ and~Ranjan~Maitra
  \IEEEcompsocitemizethanks{\IEEEcompsocthanksitem The authirs are
    with the Department of Statistics at Iowa State University. Email: \{berryni,maitra\}@iastate.edu.}
    \thanks{R. Maitra's research was supported in part by the USDA
      National Institute of Food and Agriculture, Hatch project
      IOW03617.}

}
\markboth{
}%
{Berry and Maitra \MakeLowercase{\textit{et al.}}:$K$-means for skewed  groups}
\clearpage
\setcounter{page}{1}


\IEEEcompsoctitleabstractindextext{%

\begin{abstract}
The $K$-means algorithm is extended to allow for partitioning of
skewed groups. Our algorithm is called TiK-Means and contributes a 
$K$-means type algorithm that assigns observations to groups while
estimating their skewness-transformation parameters. The 
resulting groups and transformation reveal general-structured clusters
that can be explained by inverting the estimated
transformation. Further, a modification of the jump statistic chooses the number of groups. Our algorithm is evaluated on simulated and
real-life datasets and then applied to a long-standing
astronomical dispute regarding the distinct kinds of gamma ray bursts. 
\end{abstract}

}
\maketitle

\IEEEdisplaynotcompsoctitleabstractindextext

\section{Introduction}
Clustering observations into distinct groups of
homogeneous observations
~\citep{murtagh_multi-dimensional_1985,ramey_nonparametric_1985,mclachlan_mixture_1988,kaufman_finding_1990,everitt_cluster_2001,fraley_model-based_2002,tibshirani_cluster_2005, kettenring_practice_2006, xu_clustering_2009} is important  for many  
applications~{\em e.g.} taxonomical
classification~\citep{michener_quantitative_1957}, market 
segmentation~\citep{hinneburg_cluster_1999}, color quantization of
images~\citep{celebi13,maitra_bootstrapping_2012}, and software
management~\citep{maitra_clustering_2001}. The task is generally
challenging with many
heuristic~\citep{johnson_hierarchical_1967,everitt_cluster_2001,jain_algorithms_1988,forgy_cluster_1965,macqueen_methods_1967,kaufman_finding_1990}
or more formal
statistical-model-based~\citep{titterington_statistical_1985,mclachlan_finite_2000,fraley_model-based_2002,melnykov_finite_2010}
approaches. However, the
$K$-means~\citep{macqueen_methods_1967,lloyd_least_1982,hartigan_algorithm_1979}
algorithm that finds partitions locally minimizing 
the within-sums-of-squares~(WSS) is the most commonly used
method. This algorithm depends on good 
initialization~\citep{maitra_initializing_2009,celebi13} and 
 ideally suited to find homogeneous spherically-dispersed
 groups. Further, $K$-means itself does not provide the 
number of groups $K$, for which many methods~\citep{krzanowski_criterion_1985,
  milligan_examination_1985,kaufman_finding_1990,hamerlyandelkan03,
  pellegandmoore00, sugar_finding_2003, maitra_bootstrapping_2012}
exist. Nevertheless, it is used (in fact, more commonly abused)
extensively because of its computational speed and simplicity.

Many adaptations of $K$-means exist. The algorithm was recently
adapted to partially-observed
datasets~\citep{chietal16,lithioandmaitra18}. 
The $K$-medoids algorithm~\citep{kaufmanandrousseeuw87,parkandjun09} is a robust alternative that decrees 
each cluster center to be a medoid (or exemplar) that is an observation in the
dataset. The $K$-medians 
\citep{jain_algorithms_1988,bradleyetal97a} and $K$-modes
\citep{perpinanandwang13} algorithms replace the cluster means in $K$-means 
with medians and modes. A different  $K$-modes algorithm~\citep{chaturvedietal01,huang97a}
partitions categorical datasets.

At its heart, the $K$-means algorithm uses Euclidean distances
between observations to decide on those with similar
characteristics. As a result, 
it partitions well when the distinguishing groups of a dataset have
features on about the same 
scale, {\it i.e.}  when the dataset has homogeneous spherical clusters.
One common approach to accommodate this assumption involves scaling
each dimension of a dataset before applying 
$K$-means. It is easy to see that this remedy can sometimes have 
disastrous consequences, {\it e.g.} when the grouping phenomenon is
what caused the differential scaling in a dimension. An alternative
approach removes skewness by summarily transforming the entire dataset,
sometimes yielding  unusable results, as seen in the showcase
application of this paper. 

\subsection{Finding Types of Gamma Ray Bursts}
\label{intro.grb}
Gamma Ray Bursts (GRBs) are high-energy electromagnetic explosions
 from supernovas or kilonovas. Observable physical properties have
 established multiple types of
 GRBs~\citep{mazetsetal81,norrisetal84,dezalayetal92} with one group
 of researchers claiming 2
 kinds~\citep{kouveliotou_identification_1993} of GRBs and another
 claiming 3 types~\citep{horvath_further_2002,huja_comparison_2009,tarnopolski_analysis_2015,horvath_duration_2016,zitouni_statistical_2015,
  mukherjee_three_1998}. These analyses used only a few of the available 
features. The 25-year-old controversy between
2 or 3 GRB types
is perhaps academic because careful model-based clustering and variable selection
\citep{chattopadhyay_gaussian-mixture-model-based_2017,chattopadhyay_multivariate_2018,almodovar-rivera_kernel-estimated_2018}
showed all available features as necessary. These studies also found 5 kinds of GRBs. 

The BATSE 4Br catalog has 1599 
GRBs fully observed in 9 features, namely $T_{50}$, 
$T_{90}$ (measuring the time to arrival for the first 50\% and 90\% of
the flux for a burst), $F_1, F_2,F_3$, $F_4$ (the time-integrated
fluences in the $20\!-\!50$, $50\!-\!100$, $100\!-\!300$, and $>300$ keV
spectral channels), $P_{64}$, $P_{256}$ and $P_{1024}$ (the peak flux recorded in intervals of $64$, $256$ and
$1024$ milliseconds).  
The features all skew heavily to the right and 
are summarily  $\log_{10}$-transformed before clustering. 
On this dataset, $K$-means with the Jump
statistic~\citep{sugar_finding_2003} was initially shown 
\citep{chattopadhyayetal07} to find 3 groups but careful
reapplication~\citep{chattopadhyay_gaussian-mixture-model-based_2017}
found $K$ to be indeterminate. 
Model-based clustering (MBC)~\citep{chattopadhyay_gaussian-mixture-model-based_2017,chattopadhyay_multivariate_2018}
found 5  
ellipsoidally-dispersed groups in the $\log_{10}$-transformed dataset.
Rather than applying $K$-means on the $\log_{10}$-transformed
features,  
we investigate  a data-driven approach
to choose feature-specific transformations before applying
$K$-means. 

This paper incorporates dimension-specific transformations into
$K$-means. Our Transformation-Infused-$K$-means (TiK-means) algorithm
learns the transformation parameters from a dataset by massaging the
features to allow for the 
detection of skewed clusters within a $K$-means
framework. Section~\ref{methodology2} details our 
strategy that also includes initialization and a modification of the Jump
statistic \citep{sugar_finding_2003} for finding $K$. 
Section~\ref{illustration} evaluates the algorithm while
Section~\ref{grb:sec} applies it to the GRB 
dataset to find five distinct
groups~\citep{chattopadhyay_gaussian-mixture-model-based_2017,chattopadhyay_multivariate_2018}. We
conclude with some discussion. 
A supplement with sections, figures and tables, referred to here
with prefix ``S'', is also available. 

\section{Methodology}
\label{methodology2}
\subsection{Preliminaries}

\subsubsection{The $K$-means Algorithm}
\label{baseKmeans}
The $K$-means algorithm \citep{lloyd_least_1982,forgy_cluster_1965,macqueen_methods_1967,hartigan_algorithm_1979}
starts with $K$ initial cluster centers and 
alternately partitions and updates cluster means, 
continuing until no substantial change occurs,
and converging to a local optimum that minimizes the WSS. The minimization
criterion means the use of Euclidean  distance in identifying similar
groups 
and implicit assumption of homogeneous  
spherically-dispersed clusteres. Disregarding this assumption while using
$K$-means, as for the BATSE 4Br GRB dataset 
can yield unintuitive or unusable results. We propose to relax the
assumption of homogeneous spherical dispersions for the groups,
allowing for both heterogeneity and skewness. We do so by
including, within $K$-means, multivariate normalizing transformations
that we discuss next.

\subsubsection{Normalizing Transformations}
The Box-Cox  (power) transform \citep{box_analysis_1964} was
originally introduced in the context of regression to account for
responses that are skewed or not Gaussian-distributed, but
 fit model assumptions after appropriate transformation. 
The commonly-used Box-Cox transformation only applies to positive
values, even though the original suggestion \citep{box_analysis_1964}
was a 
two-parameter transform that also shifted the response to the positive
half of the real line before applying the one-parameter Box-Cox
transform. Unfortunately, this shift parameter is difficult to
optimize and basically ignored. Exponentiating the observations before
applying the one-parameter transformation~\citep{manly76} can have
very severe effects. The signed Box-Cox transform
\citep{bickelanddoksum81} multiplies the sign of an observation with
the one-parameter Box-Cox transform of its absolute value. Other power
transformations~\citep{yeo_new_2000} exist but our 
specific development uses the simpler 
Inverse Hyperbolic Sine (IHS) transformation
\citep{burbidge_alternative_1988} that has one
parameter, is a smooth function and is of the form
\begin{equation}
y_i = \Psi^{-1}_\lambda(x_i)=\frac{1}{\lambda} \log \left(\lambda x_i + \sqrt{\lambda^2 x_i^2
    + 1} \right) \text{   for } \lambda \ne 0 \label{IHS}
\end{equation}
As $\lambda \rightarrow 0$, $y_i\rightarrow x_i$, so $\Psi^{-1}_{0}(x_i) \equiv x_i$. Also, the 
transformation is defined for $\lambda\in\R$ but is symmetric around 0, so 
we only consider $\lambda\in[0,\infty)$ when 
finding the optimal transformation. \eqref{IHS} is for
univariate data: for multi-dimensional datasets, we 
simply use~\eqref{IHS} on each coordinate with dimension-specific
parameters. 

IHS transformation parameters are usually estimated by finding
the $\blambda$ that best normalizes the data. 
For us, however, it is enough to find a transformed space where the
groups satisfy the underlying assumptions of 
$K$-means ({\it i.e.}, homogeneous spherical clusters). Our algorithm,
described next, estimates the transformation in concert with $K$-means
clustering.
\subsection{The TiK-means algorithm}
We first describe our algorithm for the homogeneous case where all
groups in our dataset $\bX$ of $n$ $p$-dimensional records have the
same transformation. We then consider the non-homogeneous case with 
cluster-specific transformations. 

\subsubsection{The homogeneous case}
\label{homo}
The TiK-means algorithm is designed to follow (in large part) the
traditional $K$-means
algorithm~\citep{lloyd_least_1982,forgy_cluster_1965}, with similar
iterations, but for an additional step estimating the
transformation parameters and a modified distance metric that accounts
for the transformation. To fix ideas, we present a simple
high-level outline of our algorithm and then highlight the differences
between $K$-means and our modification.
\begin{enumerate}
\item Provide initial values for the $K$ centers and 
  $\blambda$. 
\item Repeat as long as $\blambda$ or cluster assignments change:
\begin{enumerate}
\item Step $\blambda$ in the direction most improving clustering. \label{2c} 
\item Assign each observation to the group with closest mean (as per distance
  in transformed space). \label{2a}  
\item Update the group centers to be the mean of the transformed
  observations in that group. \label{2b} 
\end{enumerate}
\end{enumerate}

Beyond the usual parameters governing $K$-means, TiK-means also has a
parameter vector of interest $\blambda$ that takes values from a 
(for convenience) discrete set $\bLambda_{p}$. An element of $\blambda$, $\lambda_i$, denotes the scalar transformation parameter for dimension $i$. Also, Step \ref{2a} is
the same as in  $K$-means, except that that the dataset is
transformed using the current $\blambda$ before using 
Euclidean distance. Step \ref{2b} 
calculates group means in the transformed space.

The most substantial difference of TiK-means over $K$-means is in Step
\ref{2c}, where $\blambda = ((\lambda_i))_{i=1,2,\ldots,p}$. For
practical reasons, we choose the set, $\bLambda_i$, of values for
$\lambda_i$ to be discrete, and decide on updating $\lambda_i$ by
checking one rung up and one rung down from the current value to
decide if the update improves our clustering fit.
If no eligible steps improve the current clustering for any
coordinate, then every element of 
$\blambda$ remains unchanged.

A reviewer asked about the case when changing one $\lambda_i$ does not
improve the clustering, but moving a set of two or more would. The
TiK-means default is to step only for only one dimension per
iteration, but it contains an option to allow for one step per
dimension or one step per dimension per cluster for the nonhomogeneous
case (Section \ref{hetero}). For the homogeneous algorithm these
saddle points have not been observed to occur, but allowing for more
than one step per iteration has been seen to help with finding a
global optimum in the nonhomogeneous case. Since this is not a
well-studied practice, we suggest that using the results from TiK-means with a large step type as initializations for the algorithm with one step per iteration is a safe way to proceed.

$K$-means minimizes the WSS. TiK-means uses the same metric
in the  transformed space but needs to account for the 
transformation. We use 
a standard result [see (20.20) of\citet{billingsley_probability_1986}] on the distribution of transformations of random 
vectors:
\begin{result}~\citep{billingsley_probability_1986}
For  a continuous $p$-dimensional random vector $\bY$ with probability 
density function (PDF) $f_{\bY}(\by)$, $\bX =
\bPsi(\bY)$, for $\bPsi : \R^p \rightarrow \R^p$, has PDF
$f_{\bX} (\bx) = f_{\bY}\left(\bPsi^{-1} (\bx) \right)\abs{\bJ_{{\bPsi}^{-1}}(\bx)}$, where
$\abs{\bJ_{{\bPsi}^{-1}}(\bx)}$ is the Jacobian of $\bPsi^{-1} (\bx)$. 
\label{jacob}
\end{result}
The IHS transformation exemplifies using $\bPsi^{-1}$. From
Result~\ref{jacob}, $\bX_i$ contributes  
to the loglikelihood the additional term $\log\abs{\bJ_{{\bPsi}^{-1}}(\bX_i)}$ that involves $\blambda$. 
For the case where we have transformed observations from $K$
groups, the $k$th one having mean $\bmu_k^{(\blambda)}$ and common dispersion
of $\sigma_{\blambda}^2\bI$, the optimized loglikelihood  function is,  but for additive constants
\begin{equation}
  \begin{split}
  \ell(\hat\sigma^2,\hat\blambda,\hat\bmu_1,\hat\bmu_2,& \ldots, \hat\bmu_K;\bX)\\
&  = - \frac{np}{2}\log(\text{WSS}_{\hat\blambda}) +
\sum_{i=1}^n\log|J_{\hat\blambda}(\bX_i)|.
\end{split}
\label{objf}
\end{equation}
Equation~\eqref{objf} is also the setting for hard partitioning of observations
into homogeneous Gaussian clusters in the transformed space: we
propose using its negative value as our objective function.
For the IHS transformation, $\log|J_{\hat\blambda}(\bX_i)| = -\frac{1}{2}\sum_{j = 1}^p \log(\lambda_j^2 X_{ij}^2 + 1)$, so \eqref{objf} reduces to 
\begin{equation}
\text{Obj}_{\mbox{IHS}} = \frac{np}{2} \log(\text{WSS}) +
\frac{1}{2}\sum_{i=1}^n\sum_{j=1}^p \log(\lambda_j^2 X_{ij}^2 + 1).
\label{objfun}
\end{equation}

\subsubsection{The nonhomogenous case}
\label{hetero}
Section~\ref{homo}  outlined TiK-means for dimension-specific but
cluster-agnostic transformation parameters $\blambda_i$s. A more
general version allows for cluster- and dimension-specific transformations.  This results in $K$ separate
$p$-dimensional cluster-specific $\blambda$ vectors, or a $(k\times
p)$-dimensional $\blambda$ matrix.  The objective function, using 
similar arguments as in \eqref{objfun}, is
\begin{equation}
\text{Obj}_{\mbox{IHS};\blambda} = \frac{np}{2} \log(\text{WSS}) +
\frac{1}{2}\sum_{i=1}^n\sum_{k=1}^K\zeta_{ik}\sum_{j=1}^p
\log(\lambda_{kj}^2 X_{ij}^2 + 1),
\label{objfun2}
\end{equation}
where $\lambda_{kj}$ is the transformation parameter for the $k$th
group in the $j$th dimension and $\zeta_{ik}$ is an indicator variable
that is 1 if $\bX_i$ is in the $k$th group and 0 otherwise. $\zeta_{ik}$ differentiates \eqref{objfun} from \eqref{objfun2} by dictating which of the $K$ clusters' $\blambda$ vectors should be applied to transform $X_{i}$.
Optimizing ~\eqref{objfun2} proceeds similarly as
before, but requires more calculations and iterations before
convergence. Further, initializations have a bigger role
because of the higher dimensional $\blambda$, which leads potentially to more  local, but not global, minima.

\subsection{Additional Issues}
\subsubsection{Initialization and convergence}
\label{starts}
Like  $K$-means, our TiK-means algorithm converges to a
local, but not necessarily global, optimum. Our remedy is to
run our algorithm to convergance from many  random initial values for
$\blambda$ and means in the transformed space.
The minimizer of \eqref{objfun}  or \eqref{objfun2} (as
applicable) is our TiK-means solution. Our algorithm is similar
to $K$-means, but has an 
additional layer of complexity because of the estimation of $\blambda$ at each iteration. Also,
Steps~\ref{2c}-\ref{2b} of TiK-means are inter-dependent as 
orientation of clusters at the current iteration determines the next 
$\lambda_i$ step which also depends on the current partitioning. The
additional complexity over $K$-means slows down convergence,
especially in the non-homogeneous case for large 
$\bLambda$. 
So we first obtain cluster-agnostic $\blambda$s and initialize the
non-homogeneous algorithm with these TiK-means solutions. 
Our experience shows this approach to have good convergence.

In general, the homogeneous TiK-means algorithm has never been
observed to fall into a cycle that does not converge. For complex
problems however, the nonhomogeneous TiK-means algorithm, with poorly
initialized $k\times p$-dimensional $\blambda$ sometimes finds itself
in a loop that it can not escape. This generally happens in 
sub-optimal areas of the objective function surface, and is 
addressed with another initializer such as, for instance, the
homogeneous TiK-means solution. 

In addition to using many random starting points, the $\bLambda$ grid
must be specified so that it contains the true $\blambda$ values. If
it does not, the estimated $\blambda$ may pile up near the bounds of
the grid and result in no local optimum being found. Of
course, increasing the upper bound will allow for large $\blambda$s
to be estimated. The objective function rewards spherical clusters so
the $\blambda$ values will move based on their relative size with
other dimensions' $\blambda$s. Increasing the density of $\bLambda$ at
lower values can allow for the larger $\blambda$s to be estimated more
easily. Finally, if a $\bLambda$ grid that allows for a good estimate of $\blambda$ is difficult to find, then scaling the dataset before clustering can normalize the magnitude of $\blambda$.

\subsubsection{Data Scaling and Centering}
\label{DataScaling}
Datasets with features on very different scales are often standardized prior
to using $K$-means or other clustering algorithms to dampen the
influence of high-variability coordinates. This approach can
err~\citep{kettenring_practice_2006}, for instance, when the
widely-varying values in some dimensions are the result of cluster separation, rather than genuine scaling differences.
TiK-means is more robust to scaling than $K$-means because the $\blambda$
needed to sphere groups are allowed to vary while optimizing
\eqref{objfun}. However, scaling is sometimes still operationally
beneficial because larger-scaled features may need larger-valued
$\blambda$s. While conceptually not an issue because our $\blambda$s are
allowed to vary by coordinate, specifying a grid $\bLambda$  for
choosing the appropriate $\blambda$s becomes difficult in such
situations. This affects performance when $K$ is not known {\em a
  priori} and needs to be estimated. 
We therefore recommend scaling when the variables for clustering are on
very different scales. 

The IHS transformation is symmetric about the origin so centering of
the data affects its fit. We do not center the data in our
implementation. (Centering is not an 
issue with transformations that include a location parameter, such as
the two-parameter Box-Cox or Yeo-Johnson transformations.) 

\subsubsection{Choosing $K$}
\label{ClusterSelection}
The jump statistic~\citep{sugar_finding_2003} is an
information-theoretic formalization of the  ``elbow method" that
analyzes the decrease in WSS ({\it distortion}, more precisely) with increasing $K$. 
The statistic is  $J_k = \left[\frac{1}{np}\text{WSS}_k\right]^{-\eta} - \left[\frac{1}{np}\text{WSS}_{k-1}\right]^{-\eta}$
where  $\text{WSS}_k$ is 
the WSS from the $K$-means algorithm with $k$ groups ($\text{WSS}_0 \equiv
0$). Here, $J_k$ is the improvement from $k-1$ to $k$
groups, so the optimal $K = \argmax_k J_k$. The jump statistic's
performance is impacted by the transformation power $\eta$, recommended~\citep{sugar_finding_2003} to be the number of
effective dimensions in the dataset, with the examples in that
paper~\citep{sugar_finding_2003} using  $\eta = p/2$. 

Extending the jump statistic to the TiK-means setting is not
immediate, because the WSS does not accurately represent the
distortion of the clustered data in either the transformed or the
original space. Since the $\blambda$ is learned differently for
different $K$, the data for each $K$ is in different transformed spaces.
We propose as our distortion measure 
\eqref{objfun} or \eqref{objfun2} for the homogeneous or nonhomogeneous
cases. Neither objective function necessarily increases monotonically
with $K$, but it functions more 
similarly to the  $K$-means WSS than the transformed WSS does with
respect to the behavior of the jump statistics.

Selecting the $\eta$ in  $J_k$ 
generally requires care, and this is even more needed with
TiK-means. The case of non-homogeneous, non-spherical clusters (that forms the
setting for TiK-means) is where 
the out-of-the-box $\eta=p/2$ consistently performs poorly.
Any single across-the-board prescription of $\eta$ is
unreliable, so we  calculate our $J_k$ across a range of $\eta$s
and choosing the $K$ inside the range of candidate $K$s that maximizes
the Jump statistic for the largest number of $\eta$-values. We call
the display of $\argmax_KJ_K$ against $\eta$ the {\it jump selection plot}.

\section{Performance Evaluations}
\label{illustration}
We illustrate our algorithm on a dataset with skewed groups simulated using
CARP~\citep{maitra_simulating_2010,melnykov_carp_2011}. Our 
example is a case where $K$-means performs poorly and demonstrates the
value of TiK-means. We then evaluate performance on several standard
classification datasets. Our comparisons are with $K$-means and
Gaussian MBC~\citep{fraley_mclust_2006}. 
In all cases, numerical performance evaluations are
by the Adjusted Rand Index, or ARI, which compares the estimated
clustering assignment to the true  and takes values no more than 1,
with 1 indicating perfect agreement in the partitioning and 0 being no
better than that expected from a random clustering assignment. All 
performance results are summarized in Table~\ref{table:ExampleResTable}.
\subsection{Illustrative Example} 
\begin{figure}[h]
  \centering
  \mbox{\subfloat[]{\label{fig:toya}\includegraphics[width=0.5\columnwidth]{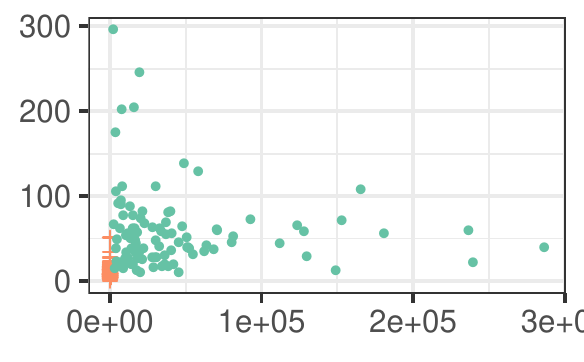}}
    \subfloat[]{\label{fig:toyb}\includegraphics[width=0.5\columnwidth]{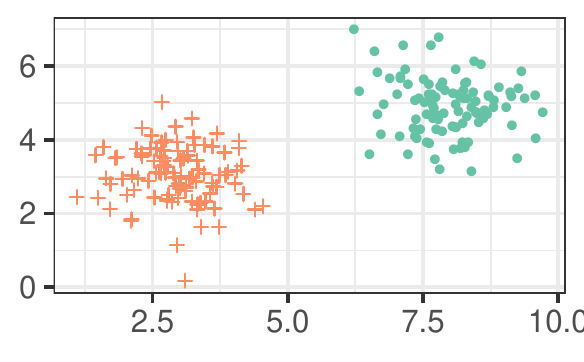}}}
  \mbox{\subfloat[]{\label{fig:toyc}\includegraphics[width=0.5\columnwidth]{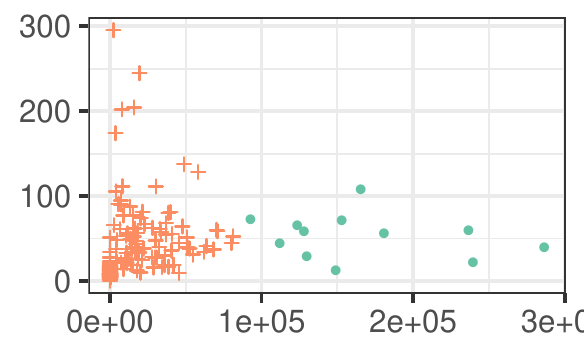}}
    \subfloat[]{\label{fig:toyd}\includegraphics[width=0.5\columnwidth]{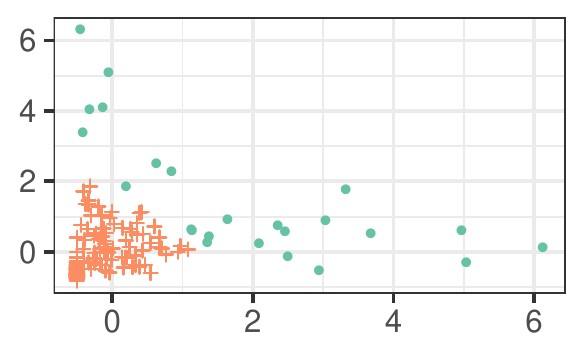}}}
    \mbox{ 
       \subfloat[]{\label{fig:toye}\includegraphics[width=0.5\columnwidth]{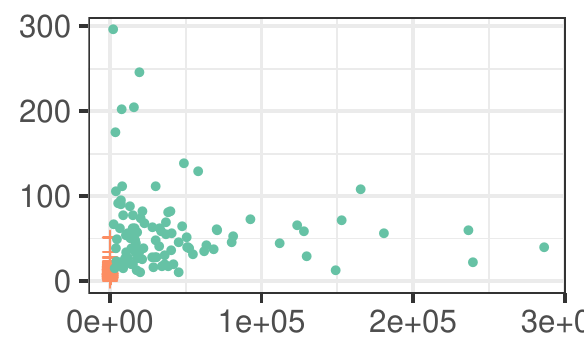}}
    \subfloat[]{\label{fig:toyf}\includegraphics[width=0.5\columnwidth]{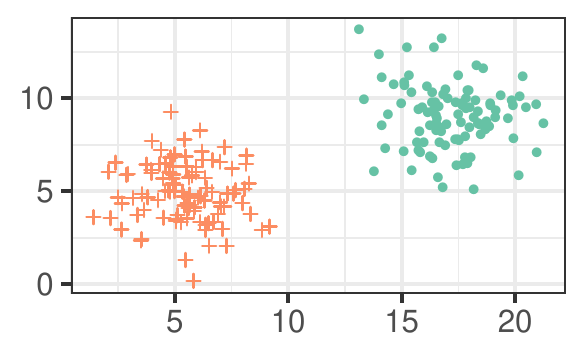}}}
  \caption[Simulated dataset for illustrating TiK-means.]{\ref{fig:toyb} shows the latent generated data. After transformation the observable data is shown in \ref{fig:toya}. Two versions of $K$-means are shown in \ref{fig:toyc} and \ref{fig:toyd}. \ref{fig:toye} shows the TiK-means results and \ref{fig:toyf} shows the inverse transformation of the data in \ref{fig:toya} estimated by TiK-means.}
    \label{fig:toyexample}
  \end{figure}

Our illustrative example uses a
two-dimensional simulated dataset of two clusters~(Figure
\ref{fig:toya}) that are easily separated in transformed
space~(Figure~\ref{fig:toyb}). This is a simulated dataset with known
groupings and IHS transformation with 
  $\blambda=\{1.4,0.9\}$.
The $K$-means algorithm with known $K$ does
poorly~(Figure~\ref{fig:toyc}), with an ARI of just
$0.013$. Clearly, the horizontal axis dominates the WSS
because of its larger scale, so that $K$-means is pretty much only
influenced by this coordinate in the clustering. Scaling the
coordinates and then applying $K$-means visually improves
performance~(Figure \ref{fig:toyd}) but still has a low ARI of
0.055. The scaling helps $K$-means use both dimensions, but is not 
enough to isolate the small-valued cluster. The original clustering is
however, perfectly~(Figure~\ref{fig:toye}) recovered by our algorithm
upon using both homogeneous and non-homogeneous transformations. 
Therefore, allowing $K$-means to transform the dataset as it assigned
clusters helped TiK-means find and isolate the distinct but  skewed groups: Figure~\ref{fig:toyf} displays the homogeneous TiK-means
solution in  terms of the back-transformed dataset and 
shows perfectly separated and homogeneous spherically-dispersed
clusters.

\begin{table*}[]
  \caption{\label{table:ExampleResTable} Performance, in terms of ARI
    of $K$-means,     MBC and TiK-means on datasets for known and
    estimated $K$.}
  \begin{tabular}{cc|c|cccc|rlrlrlrl|}
\cline{3-15}
 &  & \multicolumn{5}{c|}{Known K} & \multicolumn{8}{c|}{Estimated K} \\ \cline{3-15} 
 &  &  & \multicolumn{1}{c|}{} & \multicolumn{1}{c|}{} & \multicolumn{2}{c|}{TiK-means} & \multicolumn{2}{c|}{} & \multicolumn{2}{c|}{} & \multicolumn{4}{c|}{TiK-means} \\ \cline{6-7} \cline{12-15} 
 &  & \multirow{-2}{*}{K} & \multicolumn{1}{c|}{\multirow{-2}{*}{K-means}} & \multicolumn{1}{c|}{\multirow{-2}{*}{MBC}} & \multicolumn{1}{c|}{$\lambda_p$} & \multicolumn{1}{c|}{$\lambda_{k\times p}$} & \multicolumn{2}{c|}{\multirow{-2}{*}{K-means}} & \multicolumn{2}{c|}{\multirow{-2}{*}{MBC}} & \multicolumn{2}{c|}{$\lambda_p$} & \multicolumn{2}{c|}{$\lambda_{k\times p}$} \\ \hline
\multicolumn{1}{|c|}{} & Unscaled &  & \cellcolor[HTML]{EFEFEF}0.371 & \cellcolor[HTML]{EFEFEF}0.967 & \cellcolor[HTML]{EFEFEF}0.854 & \cellcolor[HTML]{EFEFEF}0.868 & \cellcolor[HTML]{EFEFEF}\textbf{7} & \cellcolor[HTML]{EFEFEF}0.226 & \cellcolor[HTML]{EFEFEF}\textbf{3} & \cellcolor[HTML]{EFEFEF}0.967 & \cellcolor[HTML]{EFEFEF}\textbf{3} & \cellcolor[HTML]{EFEFEF}0.854 & \cellcolor[HTML]{EFEFEF}\textbf{3} & \cellcolor[HTML]{EFEFEF}0.868 \\
\multicolumn{1}{|c|}{\multirow{-2}{*}{Wine}} & Scaled & \multirow{-2}{*}{\textbf{3}} & 0.898 & 0.930 & 0.854 & 0.933 & \textbf{7} & 0.528 & \textbf{3} & 0.930 & \textbf{3} & 0.854 & \textbf{3} & 0.933 \\ \hline
\multicolumn{1}{|c|}{} & Unscaled &  & \cellcolor[HTML]{EFEFEF}0.318 & \cellcolor[HTML]{EFEFEF}0.535 & \cellcolor[HTML]{EFEFEF}0.809 & \cellcolor[HTML]{EFEFEF}0.841 & \cellcolor[HTML]{EFEFEF}\textbf{10}  & \cellcolor[HTML]{EFEFEF}0.212 & \cellcolor[HTML]{EFEFEF}\textbf{10} & \cellcolor[HTML]{EFEFEF}0.314 & \cellcolor[HTML]{EFEFEF}\textbf{9} & \cellcolor[HTML]{EFEFEF}0.528 & \cellcolor[HTML]{EFEFEF}\textbf{9} & \cellcolor[HTML]{EFEFEF}0.570 \\
\multicolumn{1}{|c|}{} & Scaled & \multirow{-2}{*}{\textbf{3}} & 0.448 & 0.535 & 0.815 & 0.814 & \textbf{11} & 0.317 & \textbf{11} & 0.285 & \textbf{5} & 0.621 & \textbf{5} & 0.618 \\ \cline{2-15} 
\multicolumn{1}{|c|}{} & Unscaled &  & \cellcolor[HTML]{EFEFEF}0.444 & \cellcolor[HTML]{EFEFEF}0.649 & \cellcolor[HTML]{EFEFEF}0.761 & \cellcolor[HTML]{EFEFEF}0.781 & \cellcolor[HTML]{EFEFEF}\textbf{10} & \cellcolor[HTML]{EFEFEF}0.424 & \cellcolor[HTML]{EFEFEF}\textbf{10} & \cellcolor[HTML]{EFEFEF}0.591 & \cellcolor[HTML]{EFEFEF}\textbf{9} & \cellcolor[HTML]{EFEFEF}0.761 & \cellcolor[HTML]{EFEFEF}\textbf{9} & \cellcolor[HTML]{EFEFEF}0.781 \\
\multicolumn{1}{|c|}{\multirow{-4}{*}{Olive Oils}} & Scaled & \multirow{-2}{*}{\textbf{9}} & 0.632 & 0.659 & 0.629 & 0.622 & \textbf{11} & 0.542 & \textbf{11} & 0.544 & \textbf{5} & 0.780 & \textbf{5} & 0.778 \\ \hline
\multicolumn{1}{|c|}{} & Unscaled &  & \cellcolor[HTML]{EFEFEF}0.034 & \cellcolor[HTML]{EFEFEF}0.393 & \cellcolor[HTML]{EFEFEF}0.946 & \cellcolor[HTML]{EFEFEF}0.997 & \cellcolor[HTML]{EFEFEF}\textbf{5} & \cellcolor[HTML]{EFEFEF}0.118 & \cellcolor[HTML]{EFEFEF}\textbf{5} & \cellcolor[HTML]{EFEFEF}0.338 & \cellcolor[HTML]{EFEFEF}\textbf{5} & \cellcolor[HTML]{EFEFEF}0.927 & \cellcolor[HTML]{EFEFEF}\textbf{3} & \cellcolor[HTML]{EFEFEF}0.958 \\
\multicolumn{1}{|c|}{\multirow{-2}{*}{SDSS}} & Scaled & \multirow{-2}{*}{\textbf{2}} & 0.994 & 0.391 & 1 & 0.997 & \textbf{5} & 0.840 & \textbf{5} & 0.338 & \textbf{5} & 0.925 & \textbf{4} & 0.6709 \\ \hline
\multicolumn{1}{|c|}{} & Unscaled &  & \cellcolor[HTML]{EFEFEF}0.717 & \cellcolor[HTML]{EFEFEF}0.737 & \cellcolor[HTML]{EFEFEF}0.675 & \cellcolor[HTML]{EFEFEF}0.721 & \cellcolor[HTML]{EFEFEF}\textbf{7} & \cellcolor[HTML]{EFEFEF}0.448 & \cellcolor[HTML]{EFEFEF}\textbf{4} & \cellcolor[HTML]{EFEFEF}0.481 & \cellcolor[HTML]{EFEFEF}\textbf{3} & \cellcolor[HTML]{EFEFEF}0.675 & \cellcolor[HTML]{EFEFEF}\textbf{3} & \cellcolor[HTML]{EFEFEF}0.721 \\
\multicolumn{1}{|c|}{\multirow{-2}{*}{Seeds}} & Scaled & \multirow{-2}{*}{\textbf{3}} & 0.773 & 0.712 & 0.675 & 0.785 & \textbf{7} & 0.428 & \textbf{4} & 0.580 & \textbf{7} & 0.378 & \textbf{6} & 0.449 \\ \hline
\multicolumn{1}{|c|}{} & Unscaled &  & \cellcolor[HTML]{EFEFEF}0.532 & \cellcolor[HTML]{EFEFEF}0.640 & \cellcolor[HTML]{EFEFEF}0.570 & \cellcolor[HTML]{EFEFEF}-& \cellcolor[HTML]{EFEFEF}\textbf{20} & \cellcolor[HTML]{EFEFEF}0.524 & \cellcolor[HTML]{EFEFEF}\textbf{19} & \cellcolor[HTML]{EFEFEF}0.557 & \cellcolor[HTML]{EFEFEF}\textbf{13} & \cellcolor[HTML]{EFEFEF}0.639 & \multicolumn{2}{c|}{\cellcolor[HTML]{EFEFEF}-} \\
\multicolumn{1}{|c|}{\multirow{-2}{*}{Pen Digits}} & Scaled & \multirow{-2}{*}{\textbf{10}} & 0.524 & 0.633 & 0.556 & - & \textbf{20} & 0.562 & \textbf{20} & 0.480 & \textbf{8} & 0.546 & \multicolumn{2}{c|}{-} \\ \hline
\end{tabular}
\end{table*}

\subsection{Performance on Classification Datasets}
Our  evaluations are on  classification datasets with
different  $(n,p,K)$ that are popular for evaluating clustering
algorithms. Our datasets have $n$ ranging from $n=150$
to $n=10992$ and $p$ from $p=3$ to $p=16$. We set
$K_{\max}=\min(2K_{\mbox{true}}+1,20)$ when the $K$ needs to be 
estimated. In all cases, we evaluated performance on both scaled and
unscaled datasets.
\subsubsection{Wines}
\label{wineSection}
\begin{figure}[h]
  \centering
\includegraphics[width = 0.5\textwidth]{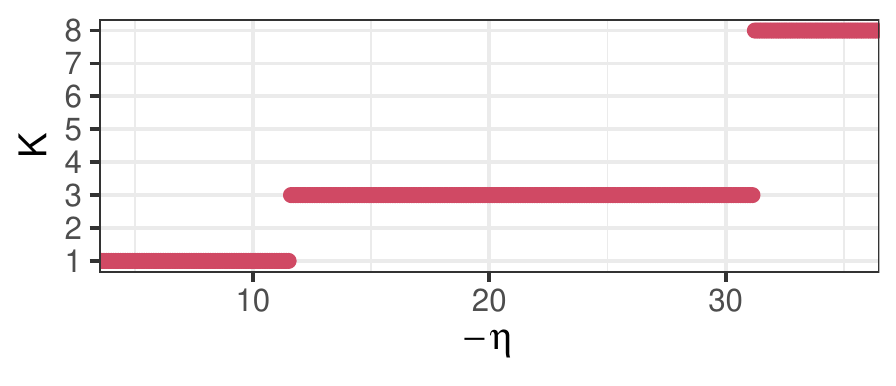}
\caption{The Jump selection plot for the wines dataset. $\hat K = 3$ is chosen, as every $\eta$ between the $\eta$s that choose $K = 1$ and the $\eta$s that choose $K=8$ has a maximum jump statistic at $3$.}
\label{fig:Wine_KSelect}
\end{figure}

This dataset~\citep{forinaetal88,aeberhardetal92} has measurements on $p=13$ chemical components
of 178 wines from $K=3$  (Barolo, Grignolino and Barbera) cultivars. TiK-means
correctly estimates $K$ (see Figure~\ref{fig:Wine_KSelect} for the
jump selection plot) with very good performance
(Table~\ref{table:ExampleResTable}) that (in the non-homogeneous case)
almost matches MBC which is known to perform well on this dataset.
However, the much-abused $K$-means algorithm, while reasonably
competitive for $K=3$ on the scaled dataset, chooses 8 groups for both
the raw and scaled versions of the dataset when $K$ is not
known. Performance is substantially poorer for such a high choice of
$K$.  However, TiK-means performs creditably and  nearly recovers the
MBC grouping.   

\subsubsection{Olive Oils}
\label{olives}
The olive oils dataset~\citep{forinaandtiscornia82,forinaetal83}
contains the concentration of 8 different fatty acids in olive oils
produced in 9 regions in 3 areas (North, South and Sardinia) of
Italy. Clustering algorithms can be evaluated at a finer
level (corresponding to regions) or a macro areal level. Observations
are on similar scales so scaling is not recommended. However, we
evaluate performance both with scaling and without scaling. 
With known $K=3$, TiK-Means recovers the true clusters better than
 $K$-means or MBC on both the scaled and unscaled
dastasets, while for known $K=9$, TiK-means is the best performer
(Table~\ref{table:ExampleResTable}) on the unscaled dataset. All
algorithms perform similarly  on the scaled dataset with
$K=9$ groups. 
Non-homogeneous TiK-means betters its homogeneous cousin 
on the scaled dataset but is similar on the unscaled dataset. 

With $K$ unknown, $K$-means and MBC both choose $K=10$ and 11
for the unscaled and scaled data. Without scaling,
TiK-means chooses $K=9$, which matches the true number of regions, but
with scaling,  TiK-means chooses $K=5$. In either case, 
TiK-means betters the other algorithms for both unscaled and scaled
datasets (Table~\ref{table:ExampleResTable}).  

\subsubsection{Sloan Digital Sky Survey}
\label{sdss}

The Sloan Digital Sky Survey (SDSS) documents physical measurements on
hundreds of millions of astronomical objects. The small subset of data
made available by \cite{wagstaff_making_2005} contains 1465 complete
records quantifying brightness, size, texture, and two versions of
shape (so $p=5$). Our objective is to differentiate galaxies from
stars. Before applying TiK-means the dataset is slightly
shifted in the texture and shape features so that each coordinate is
positive. We do so to address the issue that the IHS transformation
does not have a shift parameter and naturally splits dimensions at 0
irrespective of whether the split is coherent or not for clustering. 

Table~\ref{table:ExampleResTable} indicates that $K$ is difficult to
identify for all of the algorithms - none can correctly estimate
it. Without scaling, the nonhomogeneous TiK-means algorithm
chooses $K=3$ and gets the highest ARI of $0.958$ followed by the
homogeneous algorithm with $K=5$ and an ARI of $0.927$. On the scaled
dataset, the standard $K$-means jump statistic chooses $K=7$ and gets an
ARI of $0.830$. When $K$ is known and fixed to be $2$ the results are
more palatable. MBC unsurprisingly struggles, with the lowest ARIs
except for the unscaled $K$-means 
algorithm. On the unscaled dataset, non-homogeneous TiK-means
almost recovered the true grouping, with homogeneous
TiK-means  close behind. With scaling, homogeneous
TiK-means is perfect while non-homogeneous TiK-means and $K$-means 
performed well.  
\subsubsection{Seeds}
\label{seedsSection}
The seeds dataset \citep{charytanowicz_complete_2010} contains seven
features to differentiate 3 different kinds of wheat
kernels. On this dataset, each method performs similarly with respect to
clustering accuracy for known $K$
(Table~\ref{table:ExampleResTable}). With  unknown  $K$, only
TiK-means correctly estimates $K$ on the unscaled dataset and is the
best performer, while MBC
at $\hat K=4$ is close but performs substantially worse, and $K$-means
does even worse. All methods perform indifferently on the scaled dataset.

\subsubsection{Pen Digits}
\label{penDigitsSection}
The pen digits dataset \citep{alimoglu_combining_1996} contains $p=16$
similar-scaled features extracted from each of $n=10992$ handwritten numerical
digits. 
(Because $Kp$ is reasonably large for this dataset, we only report
homogeneous TiK-means as simultaneous optimization the $Kp$
transformation parameters over such a large-dimensional parameter
space was not successful.)
For this dataset, and with known $K=10$, scaling does not impact
performance of TiK-means, K-means or MBC. MBC is the best
performer with an ARI of 0.64 (Table~\ref{table:ExampleResTable})
while TiK-means marginally betters K-means.
\begin{figure}[!t]
\centering
\includegraphics[width = 0.5\textwidth]{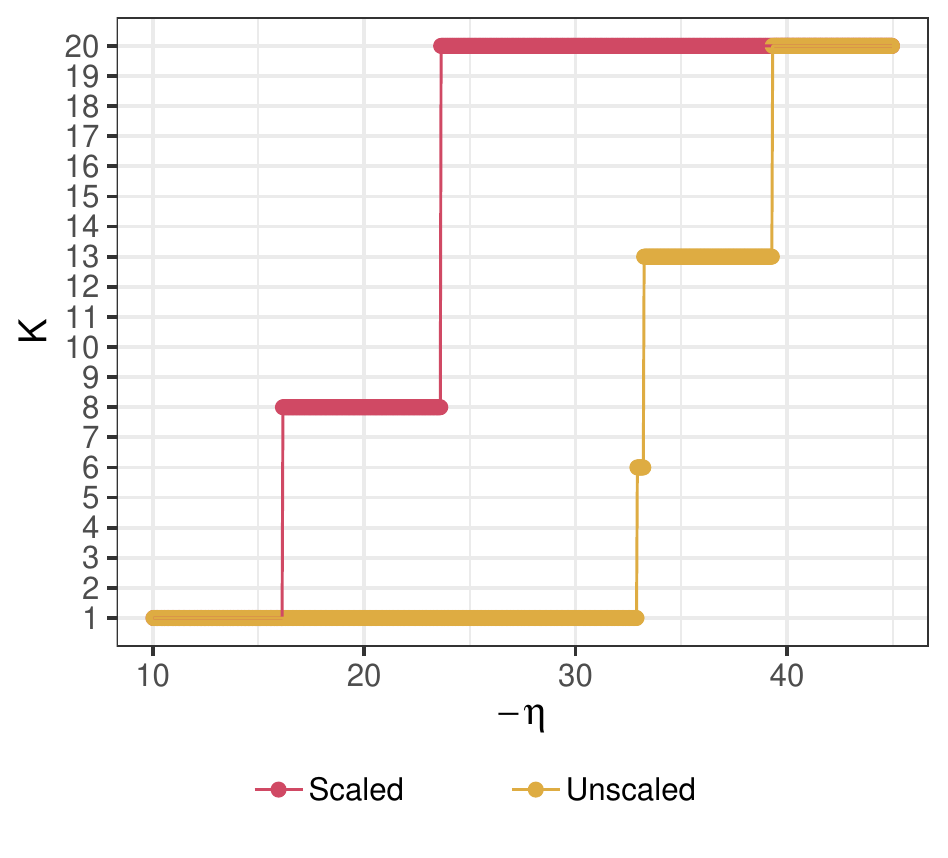}
\caption{The jump selection plot for the scaled (red) and unscaled
  (gold) pen digits datasets.}
\label{fig:digitsKSel}
\end{figure}
For unknown $K$, Figure~\ref{fig:digitsKSel} indicates that
for the scaled dataset, $K=8$ is the preference for every value of
the exponent that chooses a $K$ between $K=1$ or $K=20$.
For the unscaled
dataset, $K=13$ is estimated for all but a couple $-\eta$ values. MBC chooses $K=19$ for the unscaled dataset but is
indeterminate for the scaled version, returning the maximum
$K=20$. $K$-means with the jump
statistic \citep{sugar_finding_2003} also can not estimate $K$ for
both the scaled and unscaled datasets. In terms of performance,
TiK-means at $K=8$ is marginally beaten by $K$-means at $K=20$ for the scaled
dataset but is better than MBC. On the unscaled dataset, TiK-means is
the indisputable best performer, almost recovering in ARI the performance of
MBC with known $K=10$.



\begin{table*}[ht]
\centering
\fontsize{8.7}{10.4}\selectfont
\setlength{\tabcolsep}{1.9pt}
\caption{\label{table:penDigits_confusions}Confusion matrix of the 
  true (rows), and the groups (columns) estimated by  TiK-means
  for (a) $\hat K=8$ and (b) $\hat K=13$ for the scaled and unscaled
  datasets. The column names denote the author-specified digits that
  can be used to best characterize   the  TiK-means solutions.}
\mbox{  \subfloat[Scaled Dataset, $\hat K=8$]{%
    \label{table:k8}
\begin{tabular}{rrrrrrrrr}\hline
 & \{0\} & \{1\} & \{2\} & \{3,9\} & \{4\} & \{5,8\} & \{6\} & \{7,8\} \\ \cline{2-9}
  \multicolumn{1}{l|}{0} & 1032 &   3 &  10 &   0 &  11 &   0 &  85 &   \multicolumn{1}{r|}{2} \\ 
  \multicolumn{1}{l|}{1} &   0 & 664 & 355 & 120 &   2 &   0 &   2 &   \multicolumn{1}{r|}{0} \\ 
  \multicolumn{1}{l|}{2} &   0 &  19 & 1121 &   0 &   0 &   0 &   0 &   \multicolumn{1}{r|}{4} \\ 
  \multicolumn{1}{l|}{3} &   0 &  33 &   1 & 1020 &   1 &   0 &   0 &   \multicolumn{1}{r|}{0} \\ 
  \multicolumn{1}{l|}{4} &   0 &   9 &   1 &   6 & 1118 &   0 &  10 &   \multicolumn{1}{r|}{0} \\ 
  \multicolumn{1}{l|}{5} &   0 &   1 &   0 & 427 &   0 & 626 &   0 &   \multicolumn{1}{r|}{1} \\ 
  \multicolumn{1}{l|}{6} &   0 &   0 &  40 &   3 &   1 &   1 & 1011 &   \multicolumn{1}{r|}{0} \\ 
  \multicolumn{1}{l|}{7} &   0 & 145 &  25 &   8 &   0 &   4 &  56 & \multicolumn{1}{r|}{904} \\ 
  \multicolumn{1}{l|}{8} & 215 &  10 &  44 &  26 &   0 & 323 &  34 & \multicolumn{1}{r|}{403} \\ 
  \multicolumn{1}{l|}{9} &  17 & 243 &   0 & 672 & 122 &   0 &   0 &   \multicolumn{1}{r|}{1} \\ \cline{2-9}
\end{tabular}
}
\hspace{0.25in}
  \subfloat[Unscaled Dataset, $\hat K=13$]{%
    \label{table:k13}
\begin{tabular}{rrrrrrrrrrrrrr}\hline
 & \{0\} & \{1\} & \{2\} & \{3\} & \{4\} & \{5\} & \{6\} & \{7\} & \{8\} & \{9\} & \{0\} & \{5,9\} & \{8\} \\ \cline{2-14}
\multicolumn{1}{l|}{0} & 550 &   0 &   5 &   0 &   6 &   0 &  21 &   0 &   1 &   2 & 537 &   0 &  \multicolumn{1}{r|}{21} \\ 
\multicolumn{1}{l|}{1} &   0 & 645 & 279 &  69 &   1 &   0 &  12 &  54 &   0 &  63 &   0 &  20 &   \multicolumn{1}{r|}{0} \\ 
\multicolumn{1}{l|}{2} &   0 &  15 & 1052 &   0 &   0 &   0 &   0 &  77 &   0 &   0 &   0 &   0 &   \multicolumn{1}{r|}{0} \\ 
\multicolumn{1}{l|}{3} &   0 &  23 &   1 & 1024 &   1 &   0 &   0 &   0 &   0 &   2 &   0 &   4 &   \multicolumn{1}{r|}{0} \\ 
\multicolumn{1}{l|}{4} &   1 &  14 &   4 &   0 & 1045 &   0 &  51 &   0 &   0 &  26 &   0 &   3 &   \multicolumn{1}{r|}{0} \\ 
\multicolumn{1}{l|}{5} &   0 &   0 &   0 &  20 &   0 & 624 &  22 &   0 &   0 & 155 &   0 & 231 &   \multicolumn{1}{r|}{3} \\ 
\multicolumn{1}{l|}{6} &   0 &   0 &   1 &   0 &   3 &   1 & 1048 &   0 &   0 &   0 &   3 &   0 &   \multicolumn{1}{r|}{0} \\ 
\multicolumn{1}{l|}{7} &   0 & 149 &  19 &  74 &   1 &   4 &   1 & 891 &   2 &   0 &   0 &   0 &   \multicolumn{1}{r|}{1} \\ 
\multicolumn{1}{l|}{8} &   9 &   0 &   9 &  44 &   0 &   5 &   2 &  18 & 453 &   5 &   6 &  71 & \multicolumn{1}{r|}{433} \\ 
\multicolumn{1}{l|}{9} &  14 &  32 &   0 &  25 &  91 &   0 &   0 &   0 &   0 & 626 &   0 & 266 &   \multicolumn{1}{r|}{1} \\ 
\cline{2-14}
\end{tabular}}}
\end{table*}
Table~\ref{table:k8} shows the confusion matrix for the 
  TiK-means clustering with $8$ groups on the scaled dataset. The
column names show the digit truths that the  TiK-means clusters
seem to recover. The diagonal elements of the table show that
digits are grouped correctly together, with $0$s, $1$s,
$2$s, $3$s, $4$s, $6$s, $7$s, and $9$s grouped well, even though $3$s
and $9$s are grouped together. Many $5$s and $8$s share a cluster, but
other $5$s tend to be grouped with $3$s and $9$s, while other $8$s
are usually grouped with $7$s.
For true $K=10$, the confusion matrix
(Table~\ref{table:suppPenDigits10}) indicates not much benefit in
pre-specifying $K=10$. Though the additional clusters available to
TiK-means succeeds in separating the \{7,8\} group, $0$s are also
split into two different groups with the other extra cluster rather
than resolving the mistakes made with clustering $5$s. On the unscaled
dataset~(Table~\ref{table:k13}), and with $K=13$, every digit has a
cluster mostly to itself. The three additional groups split the $0$s
into another cluster, one splits the 8s into a new cluster, and one
contains a mixture of $5$s and $9$s. Expectedly, the common defined
clusters for  both the scaled and unscaled TiK-means results match
almost perfectly. The better performance with $K=13$ indicates a
commonly-observed trait in handwriting classification, that is, that
there is substantial variation in writing digits with more curves or edges.  

The results of our experiments show the success of TiK-means in
overcoming the homogeneous-dispersions assumption of $K$-means while
remaining within the scope of the algorithm. It performs competitively
with regards to clustering accuracy on all of the datasets presented,
sometimes performing much better than $K$-means and MBC. TiK-means is
generally as good or better at recovering the true number of clusters
as MBC, and is much better than the jump statistic with the default
transformation power. TiK-means estimates $K$ more closely when the
data is not scaled beforehand, reinforcing our earlier suggestion to
eschew scaling unless necessitated for the $\blambda$ search. We now use it
to analyze the GRB dataset. 
\section{Application to Gamma Ray Bursts dataset}
\label{grb:sec}
As discussed in Section \ref{intro.grb}, GRB datasets have so far
primarily been analyzed on the $\log_{10}$ scale to remove skewness
before clustering. This summary transformation reduces
skewness in the marginals but is somewhat arbitrary with no regard for
any potential effects on analysis. TiK-means allows us to
obtain data-driven transfomations of the variables while clustering.
In this application, the features have vastly different scales, so we
have scaled the data on the lines of the 
discussion in  Section~\ref{DataScaling} to more readily specify the
$\bLambda$ grid for estimating the IHS transformation parameters. 

\begin{figure}[!h]
\centering
\includegraphics[width = 0.5\textwidth]{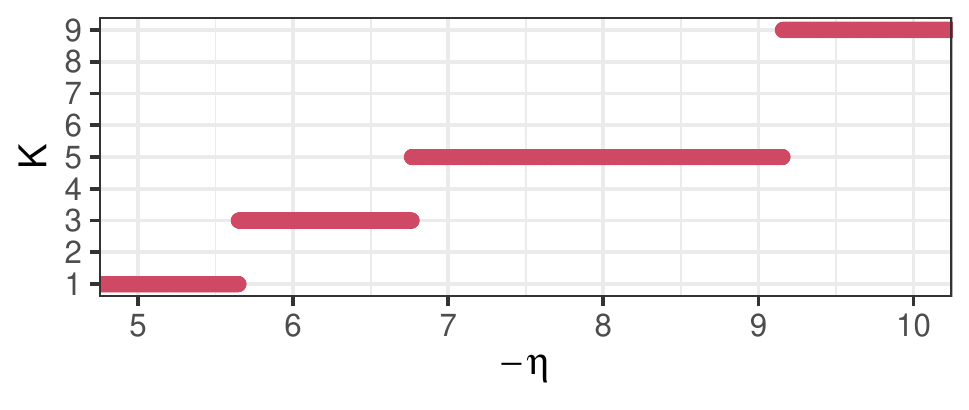}
\caption{The jump selection plot for the GRB dataset that shows
  support for 3 groups but more support for $\hat K = 5$.}
\label{fig:GRB_KSelection}
\end{figure}

Figure~\ref{fig:GRB_KSelection} displays the 
\begin{figure*}[!ht]
  \centering
  \mbox{\subfloat[]{\label{grb:radviz3d}{\animategraphics[autoplay,loop,width=0.25\textwidth]{1}{Plots/radviz3d_grb}{1}{3}}}
    \subfloat[]{\raisebox{0.175\width}{\label{table:GRBCenters}
        \begin{tabular}{r|c|ccc|c}
          \multicolumn{6}{c}{GRB Cluster Summaries} \\ \hline
          k & Size & T$_{90}$ & Fluence & Hardness$_{321}$ & Duration/Fluence/Spectrum \\
  \hline
\rowcolor{ggplot1!35}
1 & 330 & -0.278$\pm$0.55 & -29.682$\pm$1.20 & 0.446$\pm$0.02 & short/faint/soft \\ 
\rowcolor{ggplot2!35}
2 & 448 & 1.709$\pm$0.39 & -22.988$\pm$1.13 & 0.460$\pm$0.02 & long/bright/hard \\ 
\rowcolor{ggplot3!35}
3 & 190 & 1.496$\pm$0.49 & -20.576$\pm$1.61 & 0.443$\pm$0.02 & long/bright/soft \\ 
\rowcolor{ggplot4!35}
4 & 197 & 0.241$\pm$0.58 & -26.397$\pm$1.13 & 0.448$\pm$0.02 & intermediate/intermediate/soft \\ 
\rowcolor{ggplot5!35}
5 & 435 & 1.408$\pm$0.39 & -25.779$\pm$1.61 & 0.468$\pm$0.02 & long/intermediate/hard \\
    \hline
\end{tabular}%
}}}
\mbox{\subfloat[]{\label{fig:GRB.density}{\includegraphics[width =  \textwidth]{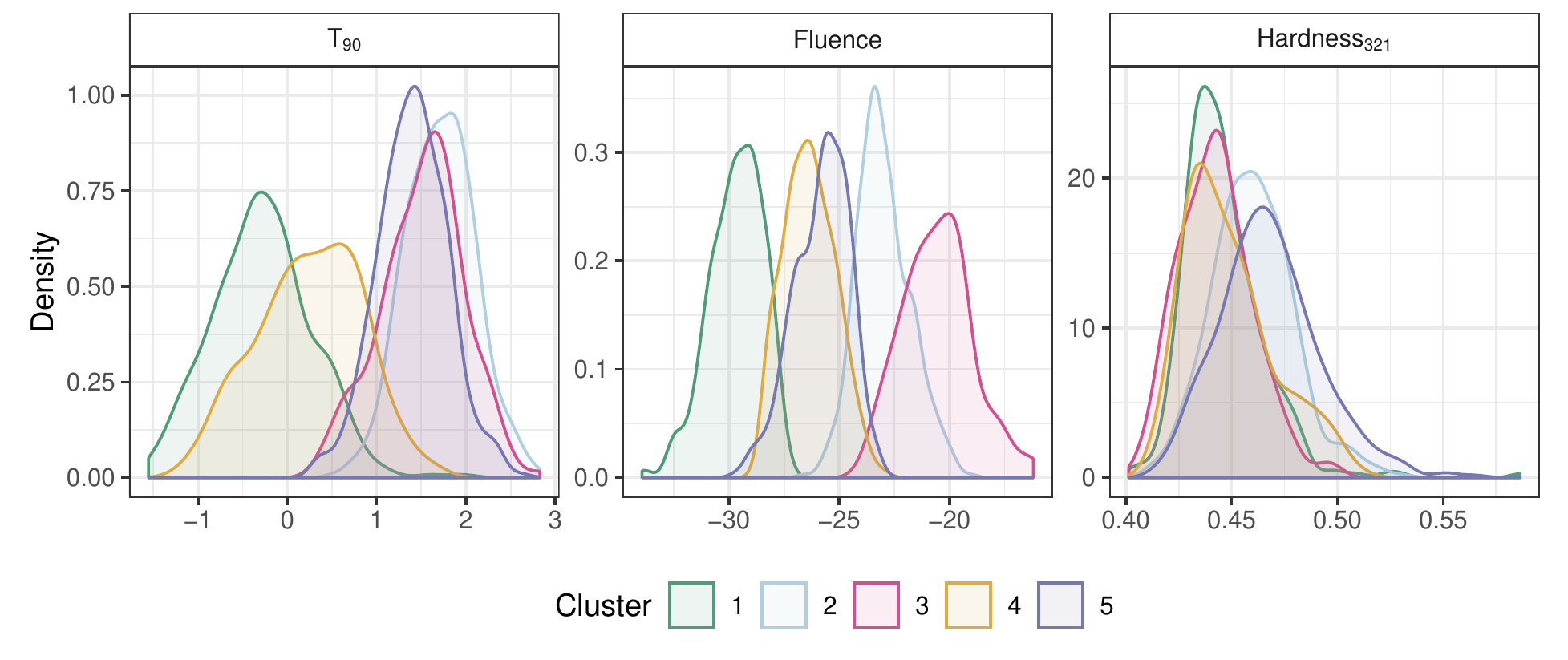}}
}}
\caption{TiK-means clustering of the GRB dataset. (a)
  Three-dimensional radial visualization plot \citep{dai_three-dimensional_2019} of obtained groups in
  transformed space. (b, c) Summary of duration
  ($T_{90}$), total fluence $F_{total}$, and spectral hardness
  $H_{321}$, introduced in \cite{mukherjee_three_1998}, for each of
  the five groups obtained using  TiK-means.}
\label{fig:GRB_Densities}
\end{figure*}
jump selection plot for $K$.
For $\eta\in(-4.31, -4.83)$, we are recommended a
3-groups solution, but 
a much larger interval for $\eta \in
(-4.83, -6.27)$, recommends 5 kinds of GRBs in the BATSE 4Br
database. So the jump selection plot suggests some
support for the 3-groups 
TiK-means solution but has a clear and overwhelming preference for the
partitioning with $K=5$.
(A similar observation holds for TiK-means on the unscaled dataset.)
Thus, 
even with $K$-means, but done on objectively-chosen transformed
spaces, our results 
demonstrate support for the presence of five distinct types of
GRBs~\citep{chattopadhyay_multivariate_2018} than the previously disputed
two-or-three types of GRBs which summarily used log-transformed
variables and used $K$-means and the Jump statistic~\citep{chattopadhyayetal07}, but which upon careful
examination~\citep{chattopadhyay_gaussian-mixture-model-based_2017} was actually indeterminate. 
Our five groups are distinct, as per the three-dimensional radial
visualization~\citep{dai_three-dimensional_2019} plot in Figure~\ref{grb:radviz3d}
which displays the most-separated projections of the transformed
dataset. We now discuss the characteristics of the obtained groups in
terms of three summary characterizations commonly used in the
astrostatistics literature~\citep{mukherjee_three_1998} to describe
the physical properties of GRBs.
These characterizations are the duration $(T_{90})$, total fluence
$(F_{total} = F_1 + F_2 + F_3 + F_4)$, and spectral hardness
$H_{321} = {F_3}/({F_1 + F_2})$. 

Figure~\ref{fig:GRB_Densities} summarizes
performance in terms of these three summary statistics. We use the
traditional $\log_{10}$ scale for these summaries in describing our
groups in order to preserve their physical interpretations.
The burst duration variable $T_{90}$ shows three groups clumped in the
long duration range, and an intermediate and a short duration cluster
that has  some separation from the large group.
The total fluence variable shows  5 homogeneous and approximately
equally-spaced groups with fairly good separation between each
other. The spectral hardness shows less separation than the other
values, but has a soft group and a hard group that are partially
separated. 
The three summary statistics characterize our groups in
terms of the paradigm of duration/fluence/spectrum as follows:
(1) long/intermediate/hard, (2) intermediate/intermediate/soft, (3)
short/faint/soft, (4) long/bright/hard, and (5)
long/bright/soft. The fourth and fifth groups  have high
fluence, but the latter is brighter and could belong to its
own \emph{very} bright fluence class. 

We end our discussion here by noting that Figure~\ref{fig:GRB.density}
also provides an indication of the reason behind the controversy between
2 and 3 GRB groups in the astrophysics community. When restricting
attention only to the most commonly 
used duration variables, it is easy to see why there would be
differing interpretation between 2 or 3 groups. However, incorporating
the additional information separates out Groups 2, 4 and 5, as can be
seen from the plots of total fluence and spectral hardness. Our
TiK-means solution presents groups that are distinct,
interpretable and in line with newer results obtained using carefully
done Gaussian~\citep{chattopadhyay_gaussian-mixture-model-based_2017} or
$t$-mixtures~\citep{chattopadhyay_multivariate_2018}
model-based or nonparametric syncytial
\citep{almodovar-rivera_kernel-estimated_2018} clustering.   

\section{Conclusions}
\label{sec:conclusions}
In this paper, we present a modification of $K$-means
called the TiK-means algorithm that preserves the simplicity and
computational advantages of $K$-means while recovering 
groups of varying complexities. The efficient and ubiquitous $K$-means clustering algorithm has inherent underlying distributional assumptions of homogeneous 
spherically dispersed groups. Such assumptions are crucial to its good
performance and several alternatives exist, but practitioners 
routinely use it without regard to its applicability in the
given context. These assumptions are relaxed in the TiK-means framework, allowing for more robust application of the $K$-means algorithm.

We provide a modified Jump
statistic~\citep{sugar_finding_2003} to determine the number of
groups. The jump statistic is dependent on a parameter 
($\eta$) that represents the number of effective dimensions in the
dataset~\citep{sugar_finding_2003} and can greatly impact
performance. We desensitize our method to the choice of
$\eta$ by calculating our modified jump-selected $\hat K$ for a range
of $\eta$s in a display called the jump selection plot and choosing
the one that is supported by the largest range of $\eta$s. Our
algorithm is made available to the scientific community through a publicly available
R~\citep{r_core_team_r_2015} package at
\href{https://github.com/berryni/TiKmeans}{https://github.com/berryni/TiKmeans}
and performs creditably on several examples where
the groups are not homogeneous and spherically dispersed. Finally, we
weigh in on the distinct kinds of GRBs in the BATSE 4Br catalog. Most
astrophysicists have hitherto used $K$-means on a few features to 
arrive at two or three groups in findings that had split the community
between two camps, one advocating for either solution. Recent
work~\citep{chattopadhyay_gaussian-mixture-model-based_2017,chattopadhyay_multivariate_2018} showed
that the $K$-means solutions were neither tenable nor determinate in
terms of finding the kinds of GRBs and that in reality, MBC with
the Bayesian Information Criterion indicates five groups of
clusters. Our analysis of the GRB dataset using TiK-means
confirms this finding and shows five well-separated groups, shedding
more light into this long-standing question with implications on
 the origin of the universe.

There are a number of areas that could benefit from further work.  One
potential downside to our algorithm when compared to $K$-means  is the
running time which can quickly increase with $p$
and/or the fineness of the search grid for $\blambda$. 
We also need to have a better understanding of the 
convergence properties of the algorithm. While the convergence of the
algorithm has not been proven like in the case of traditional
$K$-means, for any reasonable number of starting points the algorithm
converges to consistent values of both $\blambda$ and cluster
assignments. However, it is conceivable that there exist cases where
$\blambda$ and the cluster assignments reach a point where they never
converge, and instead oscillate back and forth. We know that $K$-means
alone will converge to a local 
minimum and that a hill climbing $\blambda$ step will as well, but it
would be worth investigating if both happening simultaneously
guarantees local convergence. Additionally, the current algorithm is
implemented in conjunction with a Lloyd's
algorithm~\citep{lloyd_least_1982}. While easy to follow and modify, later algorithms do a
better job at reducing the number of computations and limiting
compute time. Attempting a
Hartigan-Wong-style algorithm~\citep{hartigan_algorithm_1979}
could speed up TiK-means. Here, care should be taken
with regards to the {\it live set} because as $\blambda$ changes from
iteration to iteration, some of the assumptions made to allow for the shortcuts in the
Hartigan-Wong algorithm may no longer be immediate and may need rework.
Increasing $n$ has a smaller
impact on running time than increasing $p$ because the expensive 
optimization step is over the $\blambda$ space and relates to $p$. Finally, the
TiK-means  algorithm has been developed and illustrated in the context
of the IHS transformation. While the performance of TiK-means using
the IHS is quite good, additional flexibility provided by, for
instance, the Yeo-Johnson and the two-parameter Box-Cox
transformations may further improve performance. Our development is
general enough to allow for such transformations, but it would be
worth evaluating their performance.
Thus, we see that notwithstanding our comprehensive approach to
finding skewed and heterogeneous-dispersed groups within the context
of $K$-means, issues meriting further attention remain.


\ifCLASSOPTIONcaptionsoff
  \newpage
\fi

\section*{Acknowledgements}
We are very grateful to the Editor and the reviewers for their
comments on an earlier version of this article  that greatly improved
its content.  
This work was supported in part by the United States Department of
Agriculture (USDA) National Institute of Food and
Agriculture, Hatch project IOW03617. 
The content of this paper is however solely the
responsibility of the authors and does not represent the
official views of NIFA or the USDA.
\bibliographystyle{IEEEtran}

\bibliography{Clustering}

\begin{thebibliography}{10}
\providecommand{\url}[1]{#1}
\csname url@samestyle\endcsname
\providecommand{\newblock}{\relax}
\providecommand{\bibinfo}[2]{#2}
\providecommand{\BIBentrySTDinterwordspacing}{\spaceskip=0pt\relax}
\providecommand{\BIBentryALTinterwordstretchfactor}{4}
\providecommand{\BIBentryALTinterwordspacing}{\spaceskip=\fontdimen2\font plus
\BIBentryALTinterwordstretchfactor\fontdimen3\font minus
  \fontdimen4\font\relax}
\providecommand{\BIBforeignlanguage}[2]{{%
\expandafter\ifx\csname l@#1\endcsname\relax
\typeout{** WARNING: IEEEtran.bst: No hyphenation pattern has been}%
\typeout{** loaded for the language `#1'. Using the pattern for}%
\typeout{** the default language instead.}%
\else
\language=\csname l@#1\endcsname
\fi
#2}}
\providecommand{\BIBdecl}{\relax}
\BIBdecl

\bibitem{murtagh_multi-dimensional_1985}
F.~Murtagh, \emph{Multi-Dimensional Clustering Algorithms}.\hskip 1em plus
  0.5em minus 0.4em\relax Berlin; New York: {Springer-Verlag}, 1985.

\bibitem{ramey_nonparametric_1985}
D.~B. Ramey, ``Nonparametric clustering techniques,'' in \emph{Encyclopedia of
  {{Statistical Science}}}.\hskip 1em plus 0.5em minus 0.4em\relax New York:
  {Wiley}, 1985, vol.~6, pp. 318--319.

\bibitem{mclachlan_mixture_1988}
G.~J. McLachlan and K.~E. Basford, \emph{Mixture {{Models}}: {{Inference}} and
  {{Applications}} to {{Clustering}}}.\hskip 1em plus 0.5em minus 0.4em\relax
  New York: {Marcel Dekker}, 1988.

\bibitem{kaufman_finding_1990}
L.~Kaufman and P.~J. Rousseuw, \emph{Finding {{Groups}} in {{Data}}}.\hskip 1em
  plus 0.5em minus 0.4em\relax New York: {John Wiley \& Sons}, 1990.

\bibitem{everitt_cluster_2001}
B.~S. Everitt, S.~Landau, and M.~Leesem, \emph{Cluster {{Analysis}} (4th
  {{Ed}}.)}.\hskip 1em plus 0.5em minus 0.4em\relax New York: {Hodder Arnold},
  2001.

\bibitem{fraley_model-based_2002}
C.~Fraley and A.~E. Raftery, ``Model-{{Based Clustering}}, {{Discriminant
  Analysis}}, and {{Density Estimation}},'' \emph{Journal of the American
  Statistical Association}, vol.~97, pp. 611--631, 2002.

\bibitem{tibshirani_cluster_2005}
R.~J. Tibshirani and G.~Walther, ``Cluster validation by prediction strength,''
  \emph{Journal of Computational and Graphical Statistics}, vol.~14, no.~3, pp.
  511--528, 2005.

\bibitem{kettenring_practice_2006}
J.~R. Kettenring, ``The practice of cluster analysis,'' \emph{Journal of
  classification}, vol.~23, pp. 3--30, 2006.

\bibitem{xu_clustering_2009}
R.~Xu and D.~C. Wunsch, \emph{Clustering}.\hskip 1em plus 0.5em minus
  0.4em\relax NJ, Hoboken: {John Wiley \& Sons}, 2009.

\bibitem{michener_quantitative_1957}
C.~D. Michener and R.~R. Sokal, ``A quantitative approach to a problem in
  classification,'' \emph{Evolution}, vol.~11, pp. 130--162, 1957.

\bibitem{hinneburg_cluster_1999}
A.~Hinneburg and D.~Keim, ``Cluster discovery methods for large databases: From
  the past to the future,'' in \emph{Proceedings of the {{ACM SIGMOD
  International Conference}} on the {{Management}} of {{Data}}}, 1999.

\bibitem{celebi13}
M.~E. Celebi, H.~A. Kingravi, and P.~A. Vela, ``A comparative study of
  efficient initialization methods for the k-means clustering algorithm,''
  \emph{Expert Systems with Applications}, vol.~40, no.~1, pp. 200--210, 2013.

\bibitem{maitra_bootstrapping_2012}
R.~Maitra, V.~Melnykov, and S.~Lahiri, ``Bootstrapping for significance of
  compact clusters in multi-dimensional datasets,'' \emph{Journal of the
  American Statistical Association}, vol. 107, no. 497, pp. 378--392, 2012.

\bibitem{maitra_clustering_2001}
R.~Maitra, ``Clustering massive datasets with applications to software metrics
  and tomography,'' \emph{Technometrics}, vol.~43, no.~3, pp. 336--346, 2001.

\bibitem{johnson_hierarchical_1967}
S.~Johnson, ``Hierarchical clustering schemes,'' \emph{Psychometrika}, vol.
  32:3, pp. 241--254, 1967.

\bibitem{jain_algorithms_1988}
A.~Jain and R.~Dubes, \emph{Algorithms for Clustering Data}.\hskip 1em plus
  0.5em minus 0.4em\relax Englewood Cliffs, NJ: {Prentice Hall}, 1988.

\bibitem{forgy_cluster_1965}
E.~Forgy, ``Cluster {{Analysis}} of {{Multivariate Data}}: {{Efficiency}} vs.
  {{Interpretability}} of {{Classifications}},'' \emph{Biometrics}, vol.~21,
  pp. 768--780, 1965.

\bibitem{macqueen_methods_1967}
J.~MacQueen, ``Some methods for classification and analysis of multivariate
  observations,'' in \emph{Proceedings of the {{Fifth Berkeley Symposium}} on
  {{Mathematical Statistics}} and {{Probability}}, {{Volume}} 1:
  {{Statistics}}}.\hskip 1em plus 0.5em minus 0.4em\relax Berkeley, Calif.:
  {University of California Press}, 1967, pp. 281--297.

\bibitem{titterington_statistical_1985}
D.~Titterington, A.~Smith, and U.~Makov, \emph{Statistical {{Analysis}} of
  {{Finite Mixture Distributions}}}.\hskip 1em plus 0.5em minus 0.4em\relax
  Chichester, U.K.: {John Wiley \& Sons}, 1985.

\bibitem{mclachlan_finite_2000}
G.~McLachlan and D.~Peel, \emph{Finite {{Mixture Models}}}.\hskip 1em plus
  0.5em minus 0.4em\relax New York: {John Wiley and Sons, Inc.}, 2000.

\bibitem{melnykov_finite_2010}
V.~Melnykov and R.~Maitra, ``Finite mixture models and model-based
  clustering,'' \emph{Statistics Surveys}, vol.~4, pp. 80--116, 2010.

\bibitem{lloyd_least_1982}
S.~Lloyd, ``Least {{Squares Quantization}} in {{PCM}},'' \emph{IEEE
  Transactions on Information Theory}, vol.~28, no.~2, pp. 129--137, Mar. 1982.

\bibitem{hartigan_algorithm_1979}
J.~A. Hartigan and M.~A. Wong, ``Algorithm {{AS}} 136: {{A K}}-{{Means
  Clustering Algorithm}},'' \emph{Journal of the Royal Statistical Society.
  Series C (Applied Statistics)}, vol.~28, no.~1, pp. 100--108, 1979.

\bibitem{maitra_initializing_2009}
R.~Maitra, ``Initializing {{Partition}}-{{Optimization Algorithms}},''
  \emph{IEEE/ACM Transactions on Computational Biology and Bioinformatics},
  vol.~6, pp. 144--157, 2009.

\bibitem{krzanowski_criterion_1985}
W.~J. Krzanowski and Y.~Lai, ``A criterion for determining the number of groups
  in a data set using sum-of-squares clustering,'' \emph{Biometrics}, vol.~44,
  no.~1, pp. 23--34, 1985.

\bibitem{milligan_examination_1985}
G.~W. Milligan and M.~C. Cooper, ``An examination of procedures for etermining
  the number of clusters in a dataset,'' \emph{Psychometrika}, vol.~50, pp.
  159--179, 1985.

\bibitem{hamerlyandelkan03}
G.~Hamerly and C.~Elkan, ``Learning the k in k-means,'' in \emph{{{NIPS}}},
  vol.~3, 2003, pp. 281--288.

\bibitem{pellegandmoore00}
D.~Pelleg and A.~Moore, ``X-means: {{Extending K}}-means with {{Efficient
  Estimation}} of the {{Number}} of {{Clusters}},'' in \emph{In {{Proceedings}}
  of the 17th {{International Conf}}. on {{Machine Learning}}}.\hskip 1em plus
  0.5em minus 0.4em\relax {Morgan Kaufmann}, 2000, pp. 727--734.

\bibitem{sugar_finding_2003}
C.~A. Sugar and G.~M. James, ``Finding the number of clusters in a dataset,''
  \emph{Journal of the American Statistical Association}, vol.~98, no. 463,
  2003.

\bibitem{chietal16}
J.~T. Chi, E.~C. Chi, and R.~G. Baraniuk, ``K-{{POD}}: {{A Method}} for
  k-{{Means Clustering}} of {{Missing Data}},'' \emph{The American
  Statistician}, vol.~70, no.~1, pp. 91--99, 2016.

\bibitem{lithioandmaitra18}
A.~Lithio and R.~Maitra, ``An efficient k-means clustering algorithm for
  datasets with incomplete records,'' \emph{Statistical Analysis and Data
  Mining -- The ASA Data Science Journal}, vol.~11, pp. 296--311, 2018.

\bibitem{kaufmanandrousseeuw87}
L.~Kaufman and P.~J. Rousseeuw, ``Clustering by means of {{Medoids}},'' in
  \emph{Statistical {{Data Analysis Based}} on the
  {{L}}{$_{1}$}\textendash{{Norm}} and {{Related Methods}}}, Y.~Dodge,
  Ed.\hskip 1em plus 0.5em minus 0.4em\relax {North-Holland}, 1987, pp.
  405--416.

\bibitem{parkandjun09}
H.~Park and C.~Jun, ``A simple and fast algorithm for {{K}}-medoids
  clustering,'' \emph{Expert Systems with Applications}, vol.~36, no.~2, pp.
  3336--3341, 2009.

\bibitem{bradleyetal97a}
P.~S. Bradley, O.~L. Mangasarian, and W.~N. Street, ``Clustering via {{Concave
  Minimization}},'' in \emph{Advances in {{Neural Information Processing
  Systems}}}, M.~C. Mozer, M.~I. Jordan, and T.~Petsche, Eds., vol.~9.\hskip
  1em plus 0.5em minus 0.4em\relax Cambridge, Massachusetts: {MIT Press}, 1997,
  pp. 368--374.

\bibitem{perpinanandwang13}
M.~A. {Carreira-Perpi\~n\'an} and W.~Wang, ``The {{K}}-modes algorithm for
  clustering,'' \emph{CoRR}, vol. abs/1304.6478, 2013.

\bibitem{chaturvedietal01}
A.~D. Chaturvedi, P.~E. Green, and J.~D. Carroll, ``K-modes clustering,''
  \emph{Journal of Classification}, vol.~18, pp. 35--56, 2001.

\bibitem{huang97a}
Z.~Huang, ``Clustering large data sets with mixed numeric and categorical
  values,'' in \emph{Proceedings of the {{First Pacific Asia Knowledge
  Discovery}} and {{Data Mining Conference}}}.\hskip 1em plus 0.5em minus
  0.4em\relax Singapore: {World Scientific}, 1997, pp. 21--34.

\bibitem{mazetsetal81}
E.~P. Mazets, S.~V. Golenetskii, V.~N. Ilinskii, V.~N. Panov, R.~L. Aptekar,
  I.~A. Gurian, M.~P. Proskura, I.~A. Sokolov, Z.~I. Sokolova, and T.~V.
  Kharitonova, ``Catalog of cosmic gamma-ray bursts from the {{KONUS}}
  experiment data. {{I}}.'' \emph{Astrophysics and Space Science}, vol.~80, pp.
  3--83, Nov. 1981.

\bibitem{norrisetal84}
J.~P. Norris, T.~L. Cline, U.~D. Desai, and B.~J. Teegarden, ``Frequency of
  fast, narrow gamma-ray bursts,'' \emph{Nature}, vol. 308, p. 434, Mar. 1984.

\bibitem{dezalayetal92}
J.-P. Dezalay, C.~Barat, R.~Talon, R.~Syunyaev, O.~Terekhov, and A.~Kuznetsov,
  ``Short cosmic events - {{A}} subset of classical {{GRBs}}?'' in
  \emph{American {{Institute}} of {{Physics Conference Series}}}, ser. American
  {{Institute}} of {{Physics Conference Series}}, W.~S. Paciesas and G.~J.
  Fishman, Eds., vol. 265, 1992, pp. 304--309.

\bibitem{kouveliotou_identification_1993}
C.~Kouveliotou, C.~A. Meegan, G.~J. Fishman, N.~P. Bhat, {Michael S. Briggs},
  T.~M. Koshut, W.~S. Paciesas, and G.~N. Pendleton,
  ``\BIBforeignlanguage{en}{Identification of two classes of gamma-ray
  bursts},'' \emph{\BIBforeignlanguage{en}{The Astrophysical Journal}}, vol.
  413, pp. L101--L104, Aug. 1993.

\bibitem{horvath_further_2002}
I.~Horvath, ``\BIBforeignlanguage{en}{A further study of the {{BATSE
  Gamma}}-{{Ray Burst}} duration distribution},''
  \emph{\BIBforeignlanguage{en}{Astronomy \& Astrophysics}}, vol. 392, no.~3,
  pp. 791--793, Sep. 2002.

\bibitem{huja_comparison_2009}
D.~Huja, A.~Meszaros, and J.~Ripa, ``\BIBforeignlanguage{en}{A comparison of
  the gamma-ray bursts detected by {{BATSE}} and {{Swift}}},''
  \emph{\BIBforeignlanguage{en}{Astronomy \& Astrophysics}}, vol. 504, no.~1,
  pp. 67--71, Sep. 2009.

\bibitem{tarnopolski_analysis_2015}
M.~Tarnopolski, ``\BIBforeignlanguage{en}{Analysis of {{Fermi}} gamma-ray burst
  duration distribution},'' \emph{\BIBforeignlanguage{en}{Astronomy \&
  Astrophysics}}, vol. 581, p. A29, Sep. 2015.

\bibitem{horvath_duration_2016}
I.~Horvath and B.~G. Toth, ``\BIBforeignlanguage{en}{The duration distribution
  of {{Swift Gamma}}-{{Ray Bursts}}},''
  \emph{\BIBforeignlanguage{en}{Astrophysics and Space Science}}, vol. 361,
  no.~5, p. 155, Apr. 2016.

\bibitem{zitouni_statistical_2015}
H.~Zitouni, N.~Guessoum, W.~J. Azzam, and R.~Mochkovitch,
  ``\BIBforeignlanguage{en}{Statistical study of observed and intrinsic
  durations among {{BATSE}} and {{Swift}}/{{BAT GRBs}}},''
  \emph{\BIBforeignlanguage{en}{Astrophysics and Space Science}}, vol. 357,
  no.~1, p.~7, Apr. 2015.

\bibitem{mukherjee_three_1998}
S.~Mukherjee, E.~D. Feigelson, G.~Jogesh~Babu, F.~Murtagh, C.~Fraley, and
  A.~Raftery, ``\BIBforeignlanguage{en}{Three {{Types}} of {{Gamma}}-{{Ray
  Bursts}}},'' \emph{\BIBforeignlanguage{en}{The Astrophysical Journal}}, vol.
  508, no.~1, pp. 314--327, Nov. 1998.

\bibitem{chattopadhyay_gaussian-mixture-model-based_2017}
S.~Chattopadhyay and R.~Maitra, ``Gaussian-{{Mixture}}-{{Model}}-based
  {{Cluster Analysis Finds Five Kinds}} of {{Gamma Ray Bursts}} in the {{BATSE
  Catalog}},'' \emph{Monthly Notices of the Royal Astronomical Society}, vol.
  469, no.~3, pp. 3374--3389, Aug. 2017.

\bibitem{chattopadhyay_multivariate_2018}
------, ``Multivariate \$t\$-{{Mixtures}}-{{Model}}-based {{Cluster Analysis}}
  of {{BATSE Catalog Establishes Importance}} of {{All Observed Parameters}},
  {{Confirms Five Distinct Ellipsoidal Sub}}-populations of {{Gamma Ray
  Bursts}},'' \emph{Monthly Notices of the Royal Astronomical Society}, Jul.
  2018.

\bibitem{almodovar-rivera_kernel-estimated_2018}
I.~{Almod\'ovar-Rivera} and R.~Maitra, ``Kernel-estimated {{Nonparametric
  Overlap}}-{{Based Syncytial Clustering}},'' \emph{arXiv:1805.09505 [stat]},
  May 2018.

\bibitem{chattopadhyayetal07}
T.~Chattopadhyay, R.~Misra, A.~K. Chattopadhyay, and M.~Naskar, ``Statistical
  {{Evidence}} for {{Three Classes}} of {{Gamma}}-{{Ray Bursts}},''
  \emph{Astrophysical Journal}, vol. 667, no.~2, p. 1017, 2007.

\bibitem{box_analysis_1964}
G.~E.~P. Box and D.~R. Cox, ``An {{Analysis}} of {{Transformations}},''
  \emph{Journal of the Royal Statistical Society. Series B (Methodological)},
  vol.~26, no.~2, pp. 211--252, 1964.

\bibitem{manly76}
B.~F.~J. Manly, ``Exponential {{Data Transformations}},'' \emph{Journal of the
  Royal Statistical Society. Series D (The Statistician)}, vol.~25, pp. 37--42,
  1976.

\bibitem{bickelanddoksum81}
P.~J. Bickel and K.~A. Doksum, ``An {{Analysis}} of {{Transformations
  Revisited}},'' \emph{Journal of the American Statistical Association},
  vol.~76, pp. 296--311, 1981.

\bibitem{yeo_new_2000}
I.-K. Yeo and R.~A. Johnson, ``A {{New Family}} of {{Power Transformations}} to
  {{Improve Normality}} or {{Symmetry}},'' \emph{Biometrika}, vol.~87, no.~4,
  pp. 954--959, 2000.

\bibitem{burbidge_alternative_1988}
J.~B. Burbidge, L.~Magee, and A.~L. Robb, ``Alternative {{Transformations}} to
  {{Handle Extreme Values}} of the {{Dependent Variable}},'' \emph{Journal of
  the American Statistical Association}, vol.~83, no. 401, pp. 123--127, Mar.
  1988.

\bibitem{billingsley_probability_1986}
P.~Billingsley, \emph{Probability and Measure}, ser. Wiley Series in
  Probability and Mathematical Statistics.\hskip 1em plus 0.5em minus
  0.4em\relax {Wiley}, 1986.

\bibitem{maitra_simulating_2010}
R.~Maitra and V.~Melnykov, ``Simulating data to study performance of finite
  mixture modeling and clustering algorithms,'' \emph{Journal of Computational
  and Graphical Statistics}, vol.~19, no.~2, pp. 354--376, 2010.

\bibitem{melnykov_carp_2011}
V.~Melnykov and R.~Maitra, ``{{CARP}}: {{Software}} for {{Fishing Out Good
  Clustering Algorithms}},'' \emph{Journal of Machine Learning Research},
  vol.~12, pp. 69 -- 73, 2011.

\bibitem{fraley_mclust_2006}
C.~Fraley and A.~E. Raftery, ``{{MCLUST Version}} 3 for {{R}}: {{Normal Mixture
  Modeling}} and {{Model}}-{{Based Clustering}},'' {University of Washington,
  Department of Statistics}, Seattle, WA, Tech. Rep. 504, 2006.

\bibitem{forinaetal88}
M.~Forina, R.~Leardi, and S.~Lanteri, ``{{PARVUS}} - {{An Extendible Package}}
  for {{Data Exploration}}, {{Classification}} and {{Correlation}},'' 1988.

\bibitem{aeberhardetal92}
D.~C. S.~Aeberhard and O.~{de Vel}, ``Comparison of {{Classifiers}} in {{High
  Dimensional Settings}},'' {Department of Computer Science and Department of
  Mathematics and Statistics, James Cook University of North Queensland}, Tech.
  Rep. 92-02, 1992.

\bibitem{forinaandtiscornia82}
M.~Forina and E.~Tiscornia, ``Pattern recognition methods in the prediction of
  {{Italian}} olive oil origin by their fatty acid content,'' \emph{Annali di
  Chimica}, vol.~72, pp. 143--155, 1982.

\bibitem{forinaetal83}
M.~Forina, C.~Armanino, S.~Lanteri, and E.~Tiscornia, ``Classification of olive
  oils from their fatty acid composition,'' in \emph{Food {{Research}} and
  {{Data Analysis}}}.\hskip 1em plus 0.5em minus 0.4em\relax London: {Applied
  Science Publishers}, 1983, pp. 189--214.

\bibitem{wagstaff_making_2005}
K.~L. Wagstaff and V.~G. Laidler, ``Making the {{Most}} of {{Missing Values}}:
  {{Object Clustering}} with {{Partial Data}} in {{Astronomy}},'' in
  \emph{Astronomical {{Data Analysis Software}} and {{Systems XIV}}}, vol. 347,
  Dec. 2005, p. 172.

\bibitem{charytanowicz_complete_2010}
M.~Charytanowicz, J.~Niewczas, P.~Kulczycki, P.~A. Kowalski, S.~Lukasik, and
  S.~Zak, ``\BIBforeignlanguage{en}{Complete {{Gradient Clustering Algorithm}}
  for~{{Features Analysis}} of {{X}}-{{Ray Images}}},'' in
  \emph{\BIBforeignlanguage{en}{Information {{Technologies}} in
  {{Biomedicine}}}}, ser. Advances in {{Intelligent}} and {{Soft Computing}},
  E.~Pietka and J.~Kawa, Eds.\hskip 1em plus 0.5em minus 0.4em\relax {Springer
  Berlin Heidelberg}, 2010, pp. 15--24.

\bibitem{alimoglu_combining_1996}
F.~Alimoglu, Y.~Doc, D.~E. Alpaydin, D.~Dr, and Y.~Denizhan, \emph{Combining
  {{Multiple Classifiers For Pen}}-{{Based Handwritten Digit Recognition}}},
  1996.

\bibitem{dai_three-dimensional_2019}
F.~Dai, Y.~Zhu, and R.~Maitra, ``Three-dimensional radial visualization of
  {{High}}-dimensional {{Continuous}} or {{Discrete Datasets}},'' \emph{ArXiv
  e-prints}, Mar. 2019.

\bibitem{r_core_team_r_2015}
{R Core Team}, \emph{R: {{A Language}} and {{Environment}} for {{Statistical
  Computing}}}, Vienna, Austria, 2018.

\bibitem{anderson_irises_1935}
E.~Anderson, ``The {{Irises}} of the {{Gaspe Peninsula}},'' \emph{Bulletin of
  the American Iris Society}, vol.~59, pp. 2--5, 1935.

\bibitem{fisher_use_1936}
R.~A. Fisher, ``The {{Use}} of {{Multiple Measurements}} in {{Taxonomic
  Poblems}},'' \emph{Annals of Eugenics}, vol.~7, pp. 179--188, 1936.

\end{thebibliography}
\newpage
\renewcommand\thefigure{S-\arabic{figure}}\setcounter{figure}{0}
\renewcommand\thetable{S-\arabic{table}}
\renewcommand\thesection{S-\arabic{section}}
\renewcommand\thesubsection{S-\arabic{section}.\arabic{subsection}}
\renewcommand\theequation{S-\arabic{equation}}

\section*{Supplementary Materials}
\setlength{\tabcolsep}{1pt}
\section{Supplementary Materials}

We  include information about   TiK-means clustering results that were
not included in the paper. 

\subsection{Wines}

Confusion matrices for the Wine dataset {TiK-means} results are shown in Tables~\ref{table:suppWineRes}.
\begin{table}[ht]
  \caption{(a) TiK-means results for the scaled and unscaled   wines
    dataset (results are identical)  with a     $p$-dimensional
    $\blambda$. TiK-means results iwth the $k\times p$-dimensional
    $\blambda$ for the (b) scaled and (c) unscaled datasets.  \label{table:suppWineRes}
  }
  \centering
  \mbox{\subfloat[]{\label{table:suppWineRes_p}
      \begin{tabular}{cccc}
        \multicolumn{4}{c}{Unscaled/Scaled ($\blambda_{p}$)} \\ \hline
        & \multicolumn{3}{c}{TiK-means} \\  \cline{2-4}
        \multicolumn{1}{c|}{1} &  59 &   0 &   \multicolumn{1}{c|}{0} \\ 
        \multicolumn{1}{c|}{2} &   2 &  62 &   \multicolumn{1}{c|}{7} \\ 
        \multicolumn{1}{c|}{3} &   0 &   0 &  \multicolumn{1}{c|}{48} \\ 
        \cline{2-4}
      \end{tabular}}
    \subfloat[]{\label{table:suppWineRes_kpUn}
      \begin{tabular}{rrrr}
        \multicolumn{4}{c}{\begin{tabular}{@{}c@{}}Unscaled ($\blambda_{k\times p}$)\end{tabular}} \\ \hline
        & \multicolumn{3}{c}{TiK-means} \\  \cline{2-4}
        \multicolumn{1}{c|}{1} &  59 & 0 & \multicolumn{1}{c|}{0} \\ 
        \multicolumn{1}{c|}{2} &   2 &  63 & \multicolumn{1}{c|}{6} \\ 
        \multicolumn{1}{c|}{3} &   0 & 0 & \multicolumn{1}{c|}{48} \\ 
        \cline{2-4}
      \end{tabular}}
    \subfloat[]{\label{table:suppWineRes_kpS}
      \begin{tabular}{rrrr}
        \multicolumn{4}{c}{\begin{tabular}{@{}c@{}}Scaled ($\blambda_{k\times p}$)\end{tabular}} \\ \hline
        & \multicolumn{3}{c}{TiK-means} \\  \cline{2-4}
        \multicolumn{1}{c|}{1} & 59 & 0 & \multicolumn{1}{c|}{0} \\ 
        \multicolumn{1}{c|}{2} & 1 &  67 &   \multicolumn{1}{c|}{3} \\  
        \multicolumn{1}{c|}{3} & 0 & 0 &  \multicolumn{1}{c|}{48} \\ 
        \cline{2-4}
      \end{tabular}
    }
  }
\end{table}

\subsection{Iris}
\label{irisSection}
\subsubsection{Iris}
\label{iris}
The celebrated Iris dataset~\citep{anderson_irises_1935,fisher_use_1936} has 
lengths and widths of petals and sepals on 50 observations each from
three Iris' species: {\it I. setosa}, {\it I. virginica} and {\it
  I. versicolor}.  The first species is very different from the others
that are less differentiated in terms of sepal 
and petal lengths and widths. At the true $K$, MBC has the
highest ARI at $0.904$, 
with 5  misclassifications. Non-homogeneous TiK-means is close,
with 6 misclassifications and an ARI of $0.886$. Homogeneous
TiK-means has 8 misclassifications and an ARI of $0.851$. $K$-means
performs poorly on both the scaled and unscaled data, although if
the Iris dataset is scaled and not centered {\it a priori}, then it
matches the performance of non-homogeneous TiK-means. Despite this
improvement, there is 
little reason to scale the dataset before $K$-means without
knowing the true cluster assignments in advance. Only nonhomogeneous
TiK-means on the scaled dataset recovers the true number of
clusters, while MBC and homogeneous TiK-means both choose 
$K=2$. $K$-means  fails to choose $K$ with
the default $\eta = 2$, although using the jump statistic with $\eta =
4$ chooses $K=3$ \cite{sugar_finding_2003}.

Despite the $p$ - dimensional version of {TiK-means} missing two data points more than the best version of the base $K$-means algorithm, {TiK-means} was run on the dataset without manipulation whereas base $K$-means only performed well on the dataset that was scaled by Root Mean Square $\left(\text{RMS}({\bx}) = \sqrt{\frac{1}{n-1}\sum_{i=1}^n x_i^2}\right)$ without centering. While manipulating a dataset before clustering is generally fine it is difficult to know exactly which way it should be scaled in the case that the true clusters are unknown, and there is little reason to scale the iris dataset by RMS unless the true cluster assignments are known.

\begin{figure}[ht!]
  \centering
  \mbox{\subfloat[]{\label{fig:IrisJumpStatistic}{\includegraphics[width = 0.5\columnwidth]{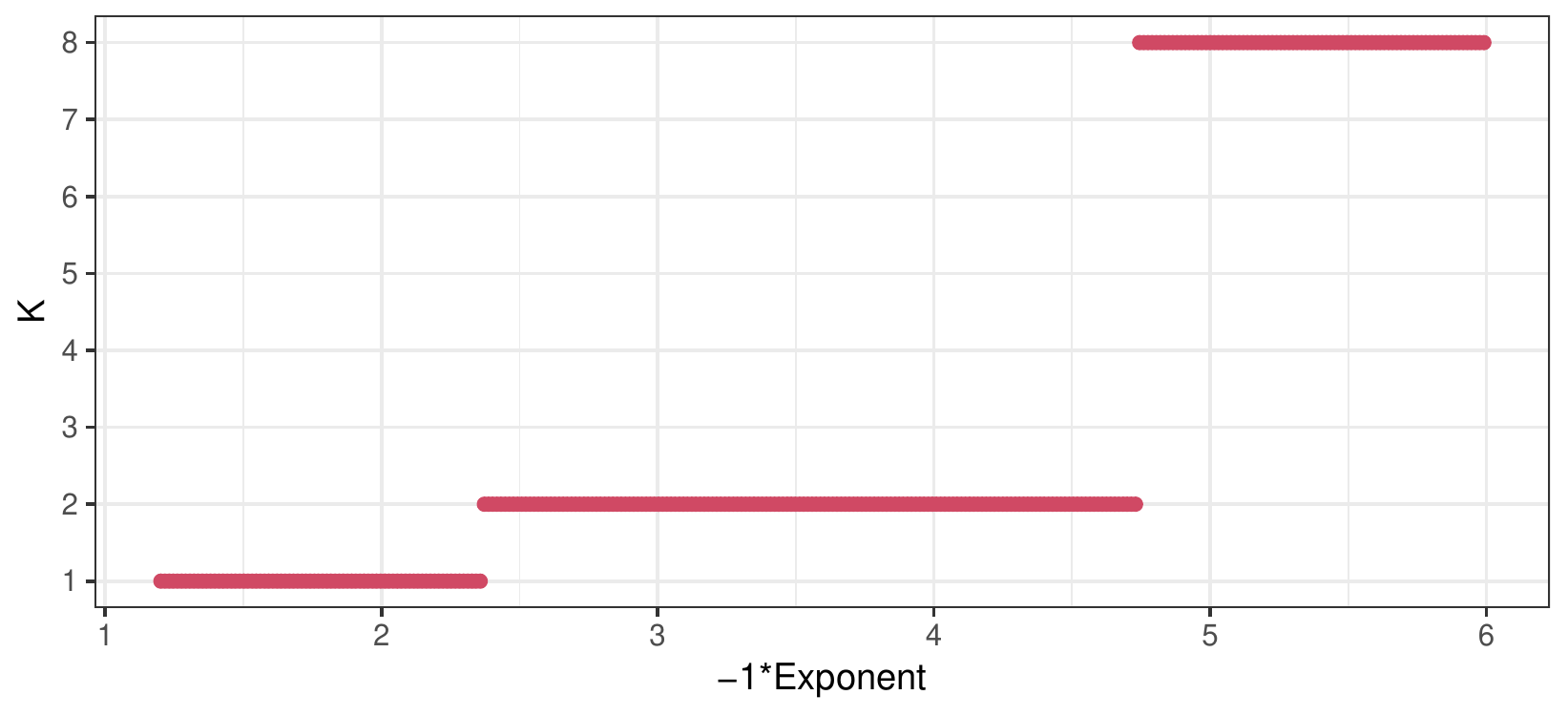}}}
    \subfloat[]{\label{fig:IrisJumpStatSlice}{\includegraphics[width = 0.5\columnwidth]{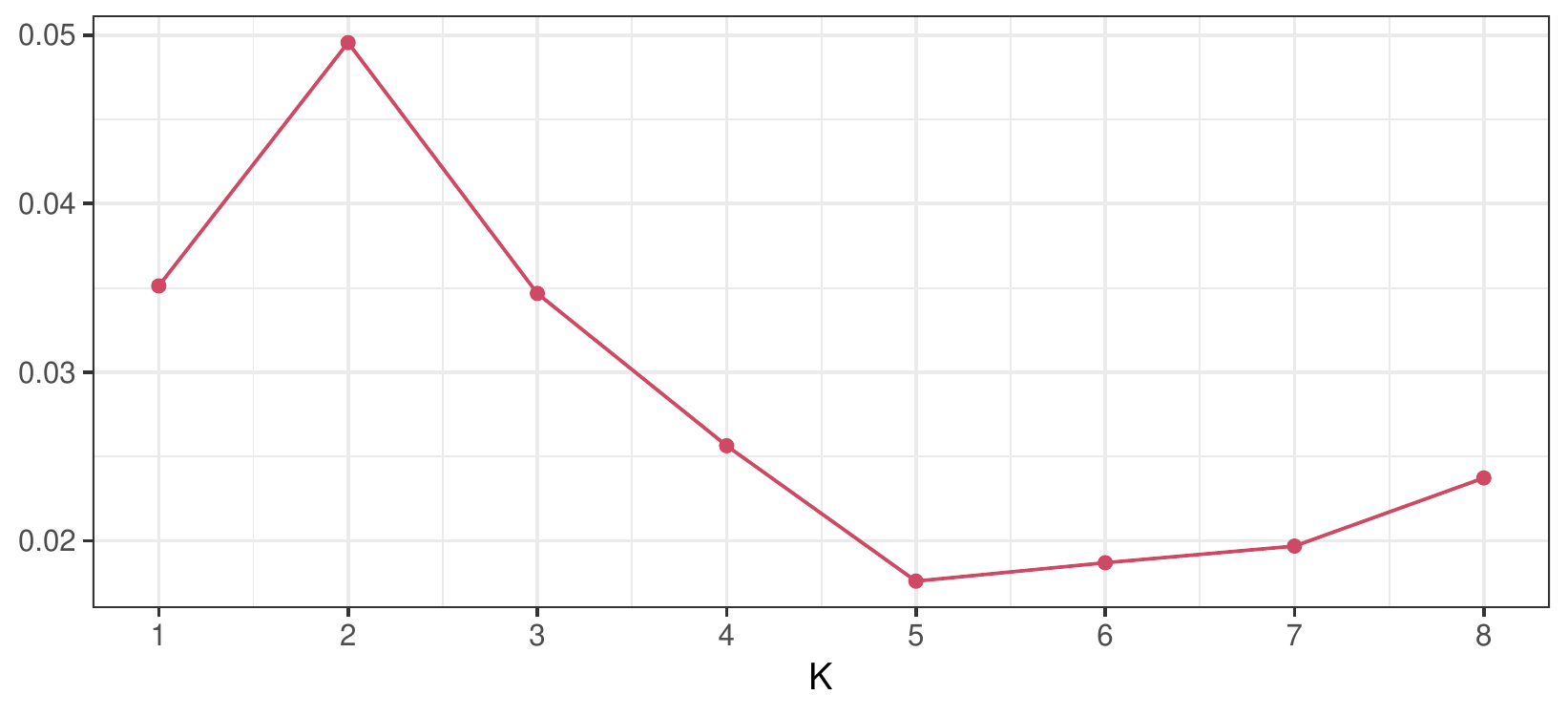}}}}
  \caption{The selection of $K$ using the jump statistic on the iris
    dataset: (a) Plot of  Jump statistic $K$  against $\eta$ and (b)
    Jump Statistic slice at $\eta = -3$. $K=2$ is the value chosen.}
  \label{fig:IrisClusterSelection}
\end{figure}

Because the labels of the iris dataset are known, it is common to cluster the dataset with 3 clusters. Without the labels it is less obvious how many clusters should be used, although 2 or 3 are the obvious answers. Using the algorithm for number of cluster selection outlined in Section \ref{ClusterSelection}, our method chooses 2 clusters as the best way to cluster the data. In Figure~\ref{fig:IrisJumpStatistic}, as you follow the dots from left to right, the selection of number of clusters goes from 1 to 2, and then to 8.

At $\eta=-3$ the method chooses $k=2$. The plot in Figure~\ref{fig:IrisJumpStatSlice} shows the jump statistics from the transformed objective functions that were chosen between. In this case $k=2$ had the largest value, followed by $k=3$. This decision between $k=2$ and $k=3$ matches our intuition for the problem. 

Tables \ref{table:suppIris_p} and \ref{table:suppIris_kp} show the confusion matrices from the {TiK-means} clustering results for $K=3$ and $p$ and $k\times p$-dimensional $\blambda$s, respectively.

\begin{table*}[ht]
  \caption{Confusion matrix for Iris dataset when clustered using   TiK-means with (a) $p$-dimensional $\blambda$ and (b) $k\times p$-dimensional $\blambda$. \label{table:suppIris}}
  \centering
  \mbox{  \subfloat[]{\label{table:suppIris_p}
      \begin{tabular}{rrrr}
        \multicolumn{4}{c}{Iris Dataset ($\blambda_p$)} \\ \hline
        & \multicolumn{3}{c}{TiK-means} \\  \cline{2-4}
        \multicolumn{1}{c|}{Setosa} & 50 & 0 & \multicolumn{1}{c|}{0} \\ 
        \multicolumn{1}{c|}{Versicolor} &   0 &  46 &   \multicolumn{1}{c|}{4} \\ 
        \multicolumn{1}{c|}{Virginica} &   0 &   4 &  \multicolumn{1}{c|}{46} \\ 
        \cline{2-4}
      \end{tabular}
    }\hspace{0.5in}
    \subfloat[]{\label{table:suppIris_kp}
      \begin{tabular}{rrrr}
        \multicolumn{4}{c}{Iris Dataset ($\blambda_{k\times p}$)} \\ \hline
        & \multicolumn{3}{c}{TiK-means} \\  \cline{2-4}
        \multicolumn{1}{c|}{Setosa} & 50 & 0 & \multicolumn{1}{c|}{0} \\ 
        \multicolumn{1}{c|}{Versicolor} &   0 &  48 &   \multicolumn{1}{c|}{2} \\ 
        \multicolumn{1}{c|}{Virginica} &   0 &   4 &  \multicolumn{1}{c|}{46} \\ 
        \cline{2-4}
      \end{tabular}
    }
  }
\end{table*}

\subsection{Olive Oils}
\subsubsection{Macro Regions $(K=3)$}
Tables \ref{table:suppOliveOil_k3} show the confusion matrices from
the  TiK-means clustering results for $K=3$ and $p$ and $k\times
p$-dimensional $\blambda$s, respectively for the scaled and unscaled datasets.
\begin{table*}[ht]
  \caption{\label{table:suppOliveOil_k3}Confusion matrix for the Olive
    Oils dataset when clustered using TiK-means with $K=3$ and (a)
    $p$-dimensional $\blambda$ (b) $k\times p$-dimensional
    $\blambda$, for the scaled dataset. Corresponding results for the
    scaled dataset are in (c) and (d). (e) Results obtained on the
    scaled (and unscaled) dataset for $\hat K=5$ groups.}
  \centering
  \mbox{\subfloat[]{\label{table:suppOliveOil_p3}
      \begin{tabular}{rrrr}
        \multicolumn{4}{c}{Olive Oils Dataset - Unscaled ($k=3$, $\blambda_p$)} \\ \hline
        & \multicolumn{3}{c}{TiK-means} \\  \cline{2-4}
        \multicolumn{1}{c|}{South} & 323 & 0 & \multicolumn{1}{c|}{0} \\ 
        \multicolumn{1}{c|}{Sardinia} &   0 &  98 &   \multicolumn{1}{c|}{0} \\ 
        \multicolumn{1}{c|}{Centre.North} &   0 & 114 &  \multicolumn{1}{c|}{37} \\ 
        \cline{2-4}
      \end{tabular}
    }
    \hspace{0.5in}
    \subfloat[]{\label{table:suppOliveOil_kp3}
      \begin{tabular}{rrrr}
        \multicolumn{4}{c}{Olive Oils Dataset - Unscaled ($k=3$, $\blambda_{p\times k}$)} \\ \hline
        & \multicolumn{3}{c}{TiK-means} \\  \cline{2-4}
        \multicolumn{1}{c|}{South} & 322 & 1 & \multicolumn{1}{c|}{0} \\ 
        \multicolumn{1}{c|}{Sardinia} &   0 &  98 &   \multicolumn{1}{c|}{0} \\ 
        \multicolumn{1}{c|}{Centre.North} &   0 & 67 &  \multicolumn{1}{c|}{84} \\ 
        \cline{2-4}
      \end{tabular}
    }
  }
  \mbox{\subfloat[]{\label{table:suppOliveOil_p3_S}
      \begin{tabular}{rrrr}
        \multicolumn{4}{c}{Olive Oils Dataset - Scaled ($k=3$, $\blambda_{p}$)} \\ \hline
        & \multicolumn{3}{c}{TiK-means} \\  \cline{2-4}
        \multicolumn{1}{c|}{South} & 323 & 0 & \multicolumn{1}{c|}{0} \\ 
        \multicolumn{1}{c|}{Sardinia} &   0 &  98 &   \multicolumn{1}{c|}{0} \\ 
        \multicolumn{1}{c|}{Centre.North} &   0 & 98 &  \multicolumn{1}{c|}{53} \\ 
        \cline{2-4}
      \end{tabular}
    }
    \hspace{0.1in}
    \subfloat[]{\label{table:suppOliveOil_kp3_S}
      \begin{tabular}{rrrr}
        \multicolumn{4}{c}{Olive Oils Dataset - Scaled ($k=3$, $\blambda_{p\times k}$)} \\ \hline
        & \multicolumn{3}{c}{TiK-means} \\  \cline{2-4}
        \multicolumn{1}{c|}{South} & 323 & 0 & \multicolumn{1}{c|}{0} \\ 
        \multicolumn{1}{c|}{Sardinia} &   0 &  98 &   \multicolumn{1}{c|}{0} \\ 
        \multicolumn{1}{c|}{Centre.North} &   0 & 100 &  \multicolumn{1}{c|}{51} \\ 
        \cline{2-4}
      \end{tabular}
    }
    \hspace{0.1in}
    \subfloat[]{\label{table:suppOliveOil_kp3_s}
      \begin{tabular}{rrrrrr}
        \multicolumn{4}{c}{Olive Oils Dataset - Scaled ($k=5$, $\blambda_{p}$)} \\ \hline
        & \multicolumn{3}{c}{TiK-means} \\  \cline{2-6}
        \multicolumn{1}{c|}{South} & 218 &   0 &   0 & 105 &  \multicolumn{1}{c|}{0} \\ 
        \multicolumn{1}{c|}{Sardinia} &   0 &  98 &   0 &   0 &  \multicolumn{1}{c|}{0} \\ 
        \multicolumn{1}{c|}{Centre.North} &   0 &   2 &  99 &   0 & \multicolumn{1}{c|}{50} \\ 
        \cline{2-6}
      \end{tabular}
    }
  }

\end{table*}

\subsubsection{Micro Regions $(K=9)$}

Tables \ref{table:suppOliveOil_k9} show the confusion matrices from the {TiK-means} clustering results for $K=9$ and $p$ and $k\times p$-dimensional $\blambda$s, respectively.

\begin{table*}[ht]
  \centering
  \caption{Confusion matrix for the Olive Oils dataset when clustered
    using TiK-means with $K=9$ and (a) $o$-dimensional $\blambda$ and
    (b) $k\times p$-dimensional
    $\blambda$. \label{table:suppOliveOil_k9} for the unscaled and (c, d)
    scaled datasets.}
  \mbox{
    \subfloat[]{\label{table:suppOliveOil_p9}
      \begin{tabular}{rrrrrrrrrr}
        \multicolumn{10}{c}{Olive Oils Dataset - Unscaled ($k=9$, $\blambda_{p}$)} \\ \hline
        & \multicolumn{9}{c}{TiK-means} \\  \cline{2-10}
        \multicolumn{1}{c|}{Apulia.north} &  23 &   2 &   0 &   0 &   0 &   0 &   0 &   0 &   \multicolumn{1}{c|}{0} \\ 
        \multicolumn{1}{c|}{Calabria} &   0 &  56 &   0 &   0 &   0 &   0 &   0 &   0 &   \multicolumn{1}{c|}{0} \\ 
        \multicolumn{1}{c|}{Apulia.south} &   0 &  13 & 193 &   0 &   0 &   0 &   0 &   0 &   \multicolumn{1}{c|}{0} \\ 
        \multicolumn{1}{c|}{Sicily} &  14 &  15 &   7 &   0 &   0 &   0 &   0 &   0 &   \multicolumn{1}{c|}{0} \\ 
        \multicolumn{1}{c|}{Sardinia.inland} &   0 &   0 &   0 &   0 &  65 &   0 &   0 &   0 &   \multicolumn{1}{c|}{0} \\ 
        \multicolumn{1}{c|}{Sardinia.coast} &   0 &   0 &   0 &   0 &  33 &   0 &   0 &   0 &   \multicolumn{1}{c|}{0} \\ 
        \multicolumn{1}{c|}{Liguria.east} &   0 &   0 &   0 &   0 &   0 &   3 &  11 &   0 &  \multicolumn{1}{c|}{36} \\ 
        \multicolumn{1}{c|}{Liguria.west} &   0 &   0 &   0 &   7 &   0 &   8 &  16 &  19 &   \multicolumn{1}{c|}{0} \\ 
        \multicolumn{1}{c|}{Umbria} &   0 &   0 &   0 &   0 &   0 &   0 &   0 &   0 &  \multicolumn{1}{c|}{51} \\  \cline{2-10}
      \end{tabular}
    }
    \hspace{0.3in}
    \subfloat[]{\label{table:suppOliveOil_kp9_u}
      \begin{tabular}{rrrrrrrrrr}
        \multicolumn{10}{c}{Olive Oils Dataset - Unscaled ($k=9$, $\blambda_{p\times k}$)} \\ \hline
        & \multicolumn{9}{c}{TiK-means} \\  \cline{2-10}
        \multicolumn{1}{c|}{Apulia.north} & 23 & 2 & 0 & 0 & 0 & 0 & 0 & 0 & \multicolumn{1}{c|}{0} \\
        \multicolumn{1}{c|}{Calabria} & 0 & 56 & 0 & 0 & 0 & 0 & 0 & 0 & \multicolumn{1}{c|}{0} \\
        \multicolumn{1}{c|}{Apulia.south} & 0 & 15 & 191 & 0 & 0 & 0 & 0 & 0 & \multicolumn{1}{c|}{0} \\
        \multicolumn{1}{c|}{Sicily} & 14 & 15 & 7 & 0 & 0 & 0 & 0 & 0 & \multicolumn{1}{c|}{0} \\
        \multicolumn{1}{c|}{Sardinia.inland} & 0 & 0 & 0 & 0 & 65 & 0 & 0 & 0 & \multicolumn{1}{c|}{0} \\
        \multicolumn{1}{c|}{Sardinia.coast} & 0 & 0 & 0 & 0 & 33 & 0 & 0 & 0 & \multicolumn{1}{c|}{0} \\
        \multicolumn{1}{c|}{Liguria.east} & 0 & 0 & 0 & 0 & 0 & 11 & 34 & 3 & \multicolumn{1}{c|}{2} \\
        \multicolumn{1}{c|}{Liguria.west} & 0 & 0 & 0 & 7 & 0 & 16 & 0 & 27 & \multicolumn{1}{c|}{0} \\
        \multicolumn{1}{c|}{Umbria} & 0 & 0 & 0 & 0 & 0 & 0 & 0 & 0 & \multicolumn{1}{c|}{51} \\
        \cline{2-10}
      \end{tabular}
    }
  }
  \mbox{\subfloat[]{\label{table:suppOliveOil_kp9}
      \begin{tabular}{rrrrrrrrrr}
        \multicolumn{10}{c}{Olive Oils Dataset - Scaled ($k=9$, $\blambda_{p}$)} \\ \hline
        & \multicolumn{9}{c}{TiK-means} \\  \cline{2-10}
        \multicolumn{1}{c|}{Apulia.north} & 23 & 2 & 0 & 0 & 0 & 0 & 0 & 0 & \multicolumn{1}{c|}{0} \\
        \multicolumn{1}{c|}{Calabria} & 0 &  54 &   2 &   0 &   0 &   0 &   0 &   0 & \multicolumn{1}{c|}{0} \\
        \multicolumn{1}{c|}{Apulia.south} & 0 &   7 & 139 &  60 &   0 &   0 &   0 &   0 & \multicolumn{1}{c|}{0} \\
        \multicolumn{1}{c|}{Sicily} & 14 &  15 &   6 &   1 &   0 &   0 &   0 &   0 & \multicolumn{1}{c|}{0} \\
        \multicolumn{1}{c|}{Sardinia.inland} & 0 &   0 &   0 &   0 &  65 &   0 &   0 &   0 & \multicolumn{1}{c|}{0} \\
        \multicolumn{1}{c|}{Sardinia.coast} & 0 &   0 &   0 &   0 &  33 &   0 &   0 &   0 & \multicolumn{1}{c|}{0} \\
        \multicolumn{1}{c|}{Liguria.east} & 0 &   0 &   0 &   0 &   0 &   0 &  43 &   3 & \multicolumn{1}{c|}{4} \\
        \multicolumn{1}{c|}{Liguria.west} & 0 &   0 &   0 &   0 &   0 &  21 &   2 &  27 & \multicolumn{1}{c|}{0} \\
        \multicolumn{1}{c|}{Umbria} & 0 &   0 &   0 &   0 &   0 &   0 &   0 &   0 & \multicolumn{1}{c|}{51} \\
        \cline{2-10}
      \end{tabular}
    }

    \hspace{0.3in}
\subfloat[]{\label{table:suppOliveOil_kp9_s}
\begin{tabular}{rrrrrrrrrr}
\multicolumn{10}{c}{Olive Oils Dataset - Scaled ($k=9$, $\blambda_{k\times p}$)} \\ \hline
 & \multicolumn{9}{c}{TiK-means} \\  \cline{2-10}
  \multicolumn{1}{c|}{Apulia.north} &  24 &   1 &   0 &   0 &   0 &   0 &   0 &   0 &   \multicolumn{1}{c|}{0} \\ 
  \multicolumn{1}{c|}{Calabria} &   1 &  54 &   1 &   0 &   0 &   0 &   0 &   0 &   \multicolumn{1}{c|}{0} \\ 
  \multicolumn{1}{c|}{Apulia.south} &   0 &   6 & 137 &  63 &   0 &   0 &   0 &   0 &   \multicolumn{1}{c|}{0} \\ 
  \multicolumn{1}{c|}{Sicily} &  17 &  14 &   3 &   2 &   0 &   0 &   0 &   0 &   \multicolumn{1}{c|}{0} \\ 
  \multicolumn{1}{c|}{Sardinia.inland} &   0 &   0 &   0 &   0 &  64 &   0 &   0 &   0 &   \multicolumn{1}{c|}{1} \\ 
  \multicolumn{1}{c|}{Sardinia.coast} &   0 &   0 &   0 &   0 &  33 &   0 &   0 &   0 &   \multicolumn{1}{c|}{0} \\ 
  \multicolumn{1}{c|}{Liguria.east} &   0 &   0 &   0 &   0 &   0 &   6 &  37 &   3 &   \multicolumn{1}{c|}{4} \\ 
  \multicolumn{1}{c|}{Liguria.west} &   0 &   0 &   0 &   0 &   0 &  23 &   0 &  27 &   \multicolumn{1}{c|}{0} \\ 
  \multicolumn{1}{c|}{Umbria} &   0 &   0 &   0 &   0 &   0 &   0 &   4 &   0 &  \multicolumn{1}{c|}{47} \\ 
\cline{2-10}
\end{tabular}
}
  }
\end{table*}

\subsection{Seeds}
Confusion matrices for the Seeds dataset using TiK-means vis-a-vis the
true grouping are in Table \ref{table:supp.seeds}.
\begin{table*}[h]
\caption{Confusion matrices for Seeds dataset when clustered with
  TiK-means at (a-c) true and (d) estimated $K$.\label{table:supp.seeds}}
\centering
\mbox{\subfloat[]{\label{table:seeds_us_3}
\begin{tabular}{rrrr}
\multicolumn{4}{c}{Seeds - Scaled/Unscaled ($K=3$, $\blambda_{p}$)} \\ \hline
 & \multicolumn{3}{c}{TiK-means} \\  \cline{2-4}
Kama &  55 &  2 &  13 \\ 
Rosa &  9 &  61 &   0 \\ 
Canadian & 2  &  0 & 68 \\ 
\cline{2-4}
\end{tabular}}
\hspace{0.5in}
\subfloat[]{\label{table:seeds_s_3}
\begin{tabular}{rrrr}
\multicolumn{4}{c}{Seeds - Scaled ($K=3$, $\blambda_{k\times p}$)} \\ \hline
 & \multicolumn{3}{c}{TiK-means} \\  \cline{2-4}
 \multicolumn{1}{c|}{Kama} &  63 &  1 &  6 \\ 
  \multicolumn{1}{c|}{Rosa} &  5 &  65 &  0 \\ 
  \multicolumn{1}{c|}{Canadian} & 4  &  0 & 66 \\ 
\cline{2-4}
\end{tabular}}
\hspace{.5in}
\subfloat[]{\begin{tabular}{rrrr}
\multicolumn{4}{c}{Seeds - Unscaled ($K=3$, $\blambda_{k\times p}$)} \\ \hline
 & \multicolumn{3}{c}{TiK-means} \\  \cline{2-4}
 \multicolumn{1}{c|}{Kama} &  58 &  1 &  11 \\ 
  \multicolumn{1}{c|}{Rosa} &  10 &  60 &  0 \\ 
  \multicolumn{1}{c|}{Canadian} & 0  &  0 &  70 \\ 
\cline{2-4}
\end{tabular}}}
\mbox{\subfloat[]{\label{table:seeds_s_7}
\begin{tabular}{rrrrrrrr}
\multicolumn{8}{c}{Seeds - Scaled ($\hat K=7$, $\blambda_{p}$)} \\ \hline
 & \multicolumn{7}{c}{TiK-means} \\  \cline{2-8}
 \multicolumn{1}{c|}{Kama} &  27 &   0 &   2 &   4 &   0 &  12 &  25 \\ 
  \multicolumn{1}{c|}{Rosa} &   0 &  30 &   0 &  17 &  21 &   0 &   2 \\ 
  \multicolumn{1}{c|}{Canadian} &   3 &   0 &  39 &   0 &   0 &  28 &0 \\ 
\cline{2-8}
\end{tabular}}
\hspace{0.5in}
\subfloat[]{\label{table:seeds_s_7u}
\begin{tabular}{rrrrrrr}
\multicolumn{7}{c}{Seeds - Scaled ($\hat K=6$, $\blambda_{k\times p}$)} \\ \hline
 & \multicolumn{6}{c}{TiK-means} \\  \cline{2-7}
 \multicolumn{1}{c|}{Kama} & 35 & 0 & 4 & 13 & 16 &  2 \\ 
  \multicolumn{1}{c|}{Rosa} & 0 & 44 & 0 & 0 & 10 &   16 \\ 
  \multicolumn{1}{c|}{Canadian} & 0 & 0 & 50 & 20 & 0 & 0 \\ 
\cline{2-7}
\end{tabular}}}
\end{table*}

\subsection{SDSS}
Confusion matrices for the SDSS dataset using TiK-means vis-a-vis the
true grouping are in Table \ref{table:supp.sdss}.

\begin{table*}[h]
\caption{Confusion matrices for the SDSS dataset when clustered with
  TiK-means with true K=2 and estimated $\hat K$. \label{table:supp.sdss}}
\centering
\mbox{\subfloat[]{\label{table:sdss_u_p}
\begin{tabular}{ccc}
\multicolumn{3}{c}{SDSS - Unscaled ($k=2$, $\blambda_{p}$)} \\ \hline
 & \multicolumn{2}{c}{TiK-means} \\  \cline{2-3}
  \multicolumn{1}{c|}{} & 0 & \multicolumn{1}{c|}{1178}\\ 
  \multicolumn{1}{c|}{} & 270 &  \multicolumn{1}{c|}{17} \\
   \cline{2-3}
\end{tabular}}
\hspace{0.1in}
\subfloat[]{\label{table:sdss_u_kp}
\begin{tabular}{ccc}
\multicolumn{3}{c}{SDSS - Unscaled ($K=2$, $\blambda_{k\times p}$)} \\ \hline
 & \multicolumn{2}{c}{TiK-means} \\  \cline{2-3}
  \multicolumn{1}{c|}{} & 1 & \multicolumn{1}{c|}{1177}\\ 
  \multicolumn{1}{c|}{} & 287 &  \multicolumn{1}{c|}{0} \\
   \cline{2-3}
\end{tabular}}
\hspace{0.1in}
\subfloat[]{\label{table:sdss_s_p}
\begin{tabular}{ccc}
\multicolumn{3}{c}{SDSS - Scaled ($K=2$, $\blambda_{p}$)} \\ \hline
 & \multicolumn{2}{c}{TiK-means} \\  \cline{2-3}
  \multicolumn{1}{c|}{} & 0 & \multicolumn{1}{c|}{1178}\\ 
  \multicolumn{1}{c|}{} & 287 &  \multicolumn{1}{c|}{0} \\
   \cline{2-3}
\end{tabular}}
\hspace{.1in}
\subfloat[]{\label{table:sdss_s_kp}
\begin{tabular}{ccc}
\multicolumn{3}{c}{SDSS - Scaled ($K=2$, $\blambda_{k\times p}$)} \\ \hline
 & \multicolumn{2}{c}{TiK-means} \\  \cline{2-3}
  \multicolumn{1}{c|}{} & 1 & \multicolumn{1}{c|}{1177}\\ 
  \multicolumn{1}{c|}{} & 287 &  \multicolumn{1}{c|}{0} \\
   \cline{2-3}
\end{tabular}}}
\mbox{\subfloat[]{\label{table:sdss_s_p5s}
\begin{tabular}{rrrrrr}
\multicolumn{6}{c}{SDSS - Scaled ($\hat K=5$, $\blambda_{p}$)} \\ \hline
 & \multicolumn{5}{c}{TiK-means} \\  \cline{2-6}
\multicolumn{1}{c|}{} &  1 & 2 & 1164 &  11 &   \multicolumn{1}{c|}{0} \\ 
\multicolumn{1}{c|}{} & 0 &   0 &   0 &  201 &  \multicolumn{1}{c|}{86} \\ 
\cline{2-6}
\end{tabular}}
\hspace{.1in}
\subfloat[]{\label{table:sdss_s_p5}
\begin{tabular}{rrrrrr}
\multicolumn{6}{c}{SDSS - Unscaled ($\hat K=5$, $\blambda_{p}$)} \\ \hline
 & \multicolumn{5}{c}{TiK-means} \\  \cline{2-6}
\multicolumn{1}{c|}{} & 3 & 2 & 0 & 1165 &  \multicolumn{1}{c|}{1} \\ 
\multicolumn{1}{c|}{} & 6 & 0 & 86 & 0 &  \multicolumn{1}{c|}{0} \\ 
\cline{2-6}
\end{tabular}}
\hspace{.1in}
\subfloat[]{\label{table:sdss_s_kp3}
\begin{tabular}{rrrrr}
\multicolumn{5}{c}{SDSS - Scaled ($\hat K=4$, $\blambda_{k\times p}$)} \\ \hline
 & \multicolumn{4}{c}{TiK-means} \\  \cline{2-5}
\multicolumn{1}{c|}{} & 31 & 0 & 1139 &  \multicolumn{1}{c|}{8} \\ 
\multicolumn{1}{c|}{} & 0 & 96 & 10 &  \multicolumn{1}{c|}{181} \\ 
\cline{2-5}
\end{tabular}} \hspace{.1in}
\subfloat[]{\label{table:sdss_s_kp4}
\begin{tabular}{rrrr}
\multicolumn{4}{c}{SDSS - Unscaled ($\hat K=3$, $\blambda_{k\times p}$)} \\ \hline
 & \multicolumn{3}{c}{TiK-means} \\  \cline{2-4}
\multicolumn{1}{c|}{} &  17 & 1161 & \multicolumn{1}{c|}{0} \\ 
\multicolumn{1}{c|}{} & 0 & 0 & \multicolumn{1}{c|}{287} \\ 
\cline{2-4}
\end{tabular}}}
\end{table*}

  \subsection{Pen Digits}
Table~\ref{table:suppPenDigits10} shows the confusion matrix of the
TiK-means solution at the true $K=10$.
  \begin{table*}[ht]
 \caption{\label{table:suppPenDigits10} Confusion matrix for the (a)
   scaled and (b) unscaled datasets showing
   differences between the true clusters, indicated by rows, and the
   {TiK-means} estimated clusters for $K=10$, indicated by
   rows. The column names denote the author specified digits that the
   TiK-means clusters seem to cover. }   \centering
 \mbox{\subfloat[]{
    \begin{tabular}{rrrrrrrrrrr}
      \multicolumn{11}{c}{Pen Digits - Scaled} \\ \hline
      & \multicolumn{10}{c}{TiK-means} \\  
      & \{0$_{i}$\} & \{1\} & \{2\} & \{3,5,9\} & \{4\} & \{5,8\} & \{6\} & \{7\} & \{8\} & \{0$_{ii}$\} \\ \cline{2-11}
      \multicolumn{1}{c|}{0} & 553 &   3 &   4 &   0 &  11 &   0 &   1 &   6 &  46 & \multicolumn{1}{c|}{519} \\ 
      \multicolumn{1}{c|}{1} &   0 & 660 & 344 & 136 &   2 &   0 &   1 &   0 &   0 &   \multicolumn{1}{c|}{0} \\ 
      \multicolumn{1}{c|}{2} &   0 &  22 & 1118 &   0 &   0 &   0 &   0 &   4 &   0 &   \multicolumn{1}{c|}{0} \\ 
      \multicolumn{1}{c|}{3} &   0 &  40 &   1 & 1014 &   0 &   0 &   0 &   0 &   0 &   \multicolumn{1}{c|}{0} \\ 
      \multicolumn{1}{c|}{4} &   0 &   9 &   1 &   5 & 1117 &   0 &  12 &   0 &   0 &   \multicolumn{1}{c|}{0} \\ 
      \multicolumn{1}{c|}{5} &   0 &   1 &   0 & 427 &   0 & 624 &   0 &   0 &   3 &   \multicolumn{1}{c|}{0} \\ 
      \multicolumn{1}{c|}{6} &   0 &   0 &  24 &   3 &   1 &   1 & 1027 &   0 &   0 &   \multicolumn{1}{c|}{0} \\ 
      \multicolumn{1}{c|}{7} &   0 & 142 &   8 &   4 &   1 &   2 &  18 & 960 &   7 &   \multicolumn{1}{c|}{0} \\ 
      \multicolumn{1}{c|}{8} &  27 &  10 &  22 &  22 &   0 & 332 &   6 &  97 & 410 & \multicolumn{1}{c|}{129} \\ 
      \multicolumn{1}{c|}{9} &   0 & 190 &   0 & 662 & 179 &   0 &   0 &   0 &   1 &  \multicolumn{1}{c|}{23} \\  \cline{2-11}
    \end{tabular}
  }}
\mbox{
  \subfloat[]{
\centering
\fontsize{8.7}{10.4}\selectfont
\begin{tabular}{rrrrrrrrrrr}
\multicolumn{11}{c}{Pen Digits - Unscaled - K = 10} \\ \hline
 & \multicolumn{10}{c}{TiK-means} \\  
 & \{0$_{i}$\} & \{1\} & \{2\} & \{3,9\} & \{4\} & \{5\} & \{6\} & \{7,8\} & \{0,8\} & \{5,9\} \\ \cline{2-11}
  \multicolumn{1}{c|}{0} & 693 &   3 &  14 &   0 &   6 &   0 &  17 &   0 & 410 &   \multicolumn{1}{c|}{0} \\ 
  \multicolumn{1}{c|}{1} &   0 & 652 & 322 &  57 &   1 &   0 &   3 &  14 &   0 & \multicolumn{1}{c|}{94} \\ 
  \multicolumn{1}{c|}{2} &   0 &  15 & 1128 &   1 &   0 &   0 &   0 &   0 &   0 &   \multicolumn{1}{c|}{0} \\ 
  \multicolumn{1}{c|}{3} &   0 &  24 &   1 & 1012 &   2 &   0 &   0 &   0 &   0 &  \multicolumn{1}{c|}{16} \\ 
  \multicolumn{1}{c|}{4} &   0 &  34 &   8 &  13 & 1046 &   0 &  23 &   0 &   0 &  \multicolumn{1}{c|}{20} \\ 
  \multicolumn{1}{c|}{5} &   0 &   0 &   0 &  94 &   0 & 627 &   1 &   0 &   0 & \multicolumn{1}{c|}{333} \\ 
  \multicolumn{1}{c|}{6} &   0 &   0 &   5 &   0 &   1 &   1 & 1044 &   0 &   0 &   \multicolumn{1}{c|}{5} \\ 
  \multicolumn{1}{c|}{7} &   0 & 156 &   9 &  30 &   0 &   4 &  49 & 894 &   0 &   \multicolumn{1}{c|}{0} \\ 
  \multicolumn{1}{c|}{8} &  18 &   1 &  18 &  12 &   0 & 126 &  45 & 345 & 432 &  \multicolumn{1}{c|}{58} \\ 
  \multicolumn{1}{c|}{9} &  16 &  70 &   0 & 598 &  98 &   0 &   0 &   1 &   0 & \multicolumn{1}{c|}{272} \\ 
  \cline{2-11}
\end{tabular}
}}
\end{table*}

\end{document}